\documentclass{article}
\usepackage{arxiv}

\usepackage{algorithm}
\usepackage{algpseudocode}
\usepackage{cite}
\usepackage{amsmath}
\usepackage{amssymb}
\usepackage{graphicx}
\usepackage{url}
\usepackage{xcolor}

\DeclareGraphicsExtensions{.png}

\definecolor{blue}{HTML}{000000}

\begin{document}

\title{\textcolor{blue}{Deep Reinforcement Learning for Stabilization\\ of Large-scale Probabilistic Boolean Networks}}

\author{Sotiris~Moschoyiannis, 
Evangelos~Chatzaroulas,
Vytenis~Sliogeris,
Yuhu~Wu
\thanks{Sotiris~Moschoyiannis, Evangelos~Chatzaroulas, and Vytenis~Sliogeris are with the School of Computer Science \& Electronic Engineering, University of Surrey, GU2 7XH, UK (e-mail: s.moschoyiannis@surrey.ac.uk). Sotiris Moschoyiannis and Vytenis Sliogeris have been partly funded by UKRI Innovate UK, grant 77032. Yuhu~Wu is with the School of Control Science and Engineering, Dalian University of Technology, 116024, Dalian, China (e-mail: wuyuhu@dlut.edu.cn).}
}

\markboth{February 2022}{}

\maketitle

\begin{abstract}
The ability to direct a Probabilistic Boolean Network (PBN) to a desired state is important to applications such as targeted therapeutics in cancer biology. \textcolor{blue}{Reinforcement Learning (RL) has been proposed as a framework that solves a discrete-time optimal control
problem cast as a Markov Decision Process. We focus on an integrative framework powered by a model-free deep RL method that can address different flavours of the control problem (e.g., with {\it or} without control inputs; attractor state {\it or} a subset of the state space as the target domain). The method is agnostic to the distribution of probabilities for the next state, hence it does not use the probability transition matrix. The time complexity is only {\it linear} on the time steps, or interactions between the agent (deep RL) and the environment (PBN), during training.
Indeed, we explore the {\it scalability}} of the deep RL approach to \textcolor{blue}{(set) stabilization} of large-scale PBNs and demonstrate successful control on large networks, including a metastatic melanoma PBN with {\it 200 nodes}. 
\end{abstract}



\section{Introduction}
\label{sec:intro}


Recent efforts to produce tools to effectively control the dynamics of complex networked systems draw from control theory, numerical methods and, more recently, network science and machine learning. A dynamical system is controllable if by intervening on the state of individual nodes the system as a whole can be driven from any initial state to a desirable state, within finite time \cite{liu-2011-controllability}. 
This notion of control finds application in  \textcolor{blue}{engineered systems but also} biological networks, \textcolor{blue}{and is often referred to as {\it stabilization}.}

Probabilistic Boolean Networks (PBNs) were introduced in \cite{schmulevich-pbn} for modelling Gene Regulatory Networks (GRNs) as complex dynamical systems. Nodes represent genes in one of two possible states of activity; 0 (not expressed), 1 (expressed). Edges indicate that genes act on each other, by means of rules represented by Boolean functions. The state of the network at each time step comprises the states of the individual nodes. PBNs extend Kauffman's Boolean Networks (BNs) \cite{kauffman-1969-metabolic} by associating more than one function with each node, one of which executes at each time step with a certain probability. This stochasticity in the network model accounts for uncertainty in gene interaction, which is inherent not only in data collection but also in cell function. Both PBNs and BNs have been extensively used to model well-known regulatory networks, see seminal work in \cite{albert-2003-topology}. 
The dynamics of PBNs adhere to Markov chain theory \cite{datta-2003-control} 
and dictate that the network, \textcolor{blue}{from any initial state, will eventually settle down to one of a limited set of steady states, the so-called {\it attractors} (fixed points or cyclic), that it cannot leave without intervention \cite{shmulevich-2002-gene-perturbation}.
}

Genes in biological systems experience sudden emergence of ordered collective behaviour, \textcolor{blue}{which corresponds to steady state behaviour in PBNs and can be modelled via attractor theory \cite{Huang-2009-cancer-attractors}}, e.g., the emergence of heterogeneous small cell lung cancer phenotypes \cite{Udyavar-2017-SCLC}. 
Therapeutic strategies may involve switching between attractors, e.g., proliferation, apoptosis in cancerous cells \cite{huang-2000-collective-behaviour}, or directing the network to a (set of) steady states that are more desirable than others, e.g., exhibit lower levels of resistance to antibiotics \cite{Reardon-2017-antibiotic-resist}. 

\textcolor{blue}{While steady states are understood in a small set of experimental GRNs in the literature, they are not generally available and it is computationally intractable to compute them in larger networks. However, attractors with larger basins of attraction tend to be more stable.  Therefore, the long-term dynamical behavior can be described by a steady-state distribution (SSD) \cite{par-2006-pbn7, kobayashi-2019-infer-pbn} which effectively captures the time spent at each network state. Hence, the control problem in large-scale PBNs takes the form of set stabilization of the network to a pre-assigned subset of network states.}  
It transpires that the ability to direct the network to a specific attractor, or a subset of network states, by means of intervention on individual nodes (genes), is central to GRNs and targeted therapeutics.

In this context, intervention at a certain time takes the form of effecting a perturbation on the state of a node, which in GRNs translates to knocking-out a gene (switch to $0$) or activating it (to $1$).
A lot of changes in cancer are of a regulatory epigenetic nature \cite{Jones-2002-epigenetic-events} and thus cells are expected to be re-configured to achieve the same outcome, \textcolor{blue}{e.g., inputs to gene regulatory elements by one signalling pathway can be substituted by another signalling pathway \cite{MAPK-nw-analysis-2022}}. Hence, cancer biology suggests that perturbations may be transient, e.g., see {single-step perturbations} in \cite{single-step-perturbations-2019}.
More concretely, this means that the network dynamics may change the state of a perturbed node in subsequent time steps.

\textcolor{blue}{Existing work in control systems engineering typically restricts perturbations to a subset of nodes, typically the {\it control nodes} of a network \cite{liu-2011-controllability,moschoyiannis-2016-control} or nodes that translate biologically, e.g., pirin and WNT5A driven induction of an invasive phenotype in melanoma cells \cite{bittner-2000-melanoma,Kim-2002-markov,par-2006-pbn7}. The general case where perturbations are considered on the full set of nodes is less studied, with the exception of \cite{CN2020-drl-pbn}, even though it is relevant in contexts where control nodes are not available, or computationally intractable to obtain.  Further, motivated by the biological properties found in various target states, different approaches perturb individual nodes' states in a PBN in order to either drive it to some attractor within a finite number of steps ({\it horizon}), or change the network's long-run behaviour by affecting its steady-state distribution (by increasing the mass probability of the target states).}

\textcolor{blue}{
Seminal work by Cheng {\it et al.} on an algebraic state space representation (ASSR) of Boolean networks, based on semi-tensor product (STP) \cite{STP-book-2011}, stimulated an avalanche of research on such networks, including controllability \cite{Liu-2015-PTM,Lu-2016-pinning,wu-2020-policy-iteration,Li-2020-outputs-pinning}.
However, ASSR linearises logical functions by enumerating their state spaces, hence it is model-based, and such methods require estimating $2^N \times 2^N$ probabilities, which quickly becomes intractable for a large number of nodes $N$.} 

\textcolor{blue}{Attempts to overcome this barrier include work on a polynomial time algorithm to identify pinned nodes \cite{polynomial-STPazuma2019,polynomial-STP-margaliot2019}, which led to subsequent developments in pinning control to rely on local neighbourhoods rather than global state information \cite{Lu-2021-acyclic}, with perturbations taking the form of deleting edges, to generate an acyclic subgraph of the network. Control of larger BNs ({\it not} PBNs which are stochastic, rather than deterministic) has been attempted in 
\cite{sensors-pin-observability-TAC-2021,distributed-pin-control-TAC-2022}. However, no guarantees are provided that the original network and its acyclic version, which is eventually controlled, have the same dynamics. In addition, there are concerns over how such changes in the network topology translate in biology since cycles are inherent in most biological pathways, e.g., see \cite{prob-inf-nw-with-cycles-breast-cancer-2019}.} 

\textcolor{blue}{The pinning control strategy has been applied to PBNs by Lin {\it et al.} \cite{stabilising-PBNs-ToC-2021} but the approach is only demonstrated on a real PBN with $N$=9 nodes, which is hardly large-scale. Moreover, the target domain is an attractor (cyclic attractor comprising 2 states) and not a subset of the state space, validated by a favourable shift in the steady state distribution (SSD) of the PBN, which is amenable to larger networks. Further, the complexity of this most recent approach is $O(N^2 +N2^d)$, hence exponential on the largest in-degree $d$ of the pinned nodes in the acyclic version of the original PBN.}

\textcolor{blue}{It transpires that the main challenge in dealing with the Boolean paradigm, as the computational counterpart of gene regulatory networks, has to do with the sheer scale of the network state space, i.e., the state transition graph or probability transition matrix, which provides a model of the system's dynamics, grows exponentially on the number of nodes. The primary objective is to develop optimal control methods that can systematically derive the series of perturbations required to direct the PBN from its current state to a target state or subset of states, thus rendering the system stabilizable at state(s) where the cell exhibits desired biological properties. In this paper we demonstrate stabilization of a Melanoma PBN within a space of $2^{200}$ states (Section \ref{sec:results-exp}).}

\textcolor{blue}{Reinforcement Learning (RL) \cite{sutton-2018-reinforcement}, by inception, addresses sequential decision-making by means of maximising a cumulative reward signal (indicator of how effective a decision was at the previous step), where the choice of action at each step influences not only the immediate reward but also subsequent states, and by virtue of that, also future rewards. Most importantly, RL provides a model-free
framework and solves a discrete-time optimal control
problem cast as an MDP.}

\textcolor{blue}{In the context of stabilization of PBNs model-free means that the distribution of probabilities of the next state, and that from each state, is not known. The RL agent learns to optimise its reward through continuous interaction with the environment - the PBN, here - from an initial state towards a terminal condition (e.g., reaching the target domain or exceeding the horizon), by following some policy on selecting actions at each state along the way (which of the $m \leq N$ nodes' state to flip). Such an {\it episodic} framework can feature in Q-Learning, which combines learning from experience (like Monte Carlo methods) {\it with} bootstrapping (like Dynamic Programming). This means there is no need to wait until the end of the episode to update the estimates of the action-values at each state (Q function). Estimates of the current step are based on estimates of the next time step, until the policy converges to the optimal policy. In this way, the PBN dynamics are learned, by means of learning the reward function.} 

\textcolor{blue}{Previous work on control that utilises  Q-Learning includes the work of Karlsen, {\it et al.} \cite{karlsen-2018-evolution} on rule-based RL, in the form of an eXtended Classifier System (XCS) which was also applied to the yeast cell cycle BN ($N$=11) in \cite{karlsen-2019-yeast}. The stabilization of PBNs with Q-Learning is studied in \cite{Acernese-2021-QLearning} but that work also only address a small apoptosis PBN ($N$=9). It transpires that Q-Learning RL struggles to converge to an optimal solution in complex and high-dimensional MDPs. }

\textcolor{blue}{Naturally, the interest shifts towards combining Q-Learning (and its model-free promise) with deep learning for scalability. Papagiannis \& Moschoyiannis in 2019 \cite{papagiannis-2019-drl-rbn} first proposed a control method based on Deep Q-Learning with experience replay, namely DDQN with Prioritized Experience Replay (PER). This was applied to control of BNs in \cite{papagiannis-2019-drl-rbn} and then to PBNs (synthetic $N$=20 and a real Melanoma $N$=7) in \cite{CN2020-drl-pbn}. Subsequently, this deep RL method was applied to solve the output tracking problem in a reduced version of the T-cell receptor kinetics model (PBCN with $N$=28) in \cite{Acernese-2020-DDQN}. Batch-mode RL has been used in \cite{sirin-2013-BatchRL} to control the Melanoma PBN ($N$=28) also studied here. Nevertheless, the advantages of combining Q-Learning with neural network function approximation to provide efficiently scalable RL control methods applicable to {\it large-scale} PBNs remain largely unexplored.}

\textcolor{blue}{In this article, we take this stream of research a step further by addressing {\it large-scale} PBNs through the application of model-free Deep Reinforcement Learning. In comparison to previous work \cite{CN2020-drl-pbn,Acernese-2020-DDQN}, we present an integrated control framework for set stabilization of large PBNs,  based on model-free deep RL (DDQN with PER) which (i) can address different flavours of the control problem, with regard to the control inputs as well as the target domain, (ii) can validate successful control in large PBNs, where computing attractors is not feasible, and (iii) has time complexity which is {\it linearly} dependent on the number of time steps and not the number of nodes in the PBN, hence a clear advancement on polynomial \cite{polynomial-STP-margaliot2019} and exponential (on largest in-degree) \cite{sensors-pin-observability-TAC-2021} complexity. 
}  

As such, the main contributions of this paper are as follows:
\begin{enumerate}
    \item \textcolor{blue}{We show that a model-free Deep RL control method for directing a PBN to a target state is {\it scalable}, with time complexity only {\it linearly} dependent on the number of time steps during training}
    \item \textcolor{blue}{We show the method to be versatile in that it can address:\\
    (a) control input nodes (when known) but also consider the full set of nodes (when not known)\\
    (b) the target domain for control to be a specific attractor (stabilization) but also a pre-assigned subset of the network state space (set stabilization)}
    \item \textcolor{blue}{We demonstrate the approach in successfully determining a control policy for PBNs and PBCNs, including stabilization of a Melanoma PBN with {\it 200 nodes}.} 
\end{enumerate}

The rest of this paper is structured as follows: Section \ref{sec:prelimin} sets out key concepts behind PBNs, \textcolor{blue}{formulates the control problem,} and outlines Deep Reinforcement Learning, focusing on DDQN with PER. The method for deriving series of perturbations ({\it control policies}) for stabilization of PBNs is developed in Section \ref{sec:ddqn-control-pbn}. The main results of applying the control method to large PBNs are presented in Section \ref{sec:results-exp}, including comparison and discussion. Finally, Section \ref{sec:concl} presents some concluding remarks and possible extensions.

\section{Preliminaries}
\label{sec:prelimin}

\subsection{Probabilistic Boolean Networks (PBNs)}
\label{sec:pbn}

PBNs are a class of discrete dynamical systems characterised by interactions over a set of $N$ nodes, each taking a Boolean value $x_i$ in $\mathcal{D}=\{0,1\}$. \textcolor{blue}{Hence, $x_i(t), i \in [1, N]$, denotes the state of the $i$-th node at time instance $t$}, and represents the expression level of the $i$-th gene in the GRN being modelled. \textcolor{blue}{The update rule of $x_i(t)$ is determined by the Boolean function $f_i([x_j(t)]_{j\in N_i})$,  and the value of $f_i$ is assigned to next state of node $x_i$,  the set $N_i \subset [1, N]$ contains the subscript indices of in-neighbours, and $f_i : \mathcal{D}^{N_i} \to \mathcal{D}$ is the logical function
chosen for node $i$ at time step $t$. The state of a PBN at time $t$, is denoted by $\mathcal{X}_t = [x_1(t), x_2(t), ...,  x_N (t)]^\top$. Then the evolution (dynamics) of the BN is represented by the following vector form:
\begin{equation}\label{BN}
\mathcal{X}_{t+1}= \left[\begin{array}{cc}
   x_1(t+1)=f_1([x_j(t)]_{j\in N_1})\\
   x_2(t+1)=f_2([x_j(t)]_{j\in  N_2})\\
   \vdots\\
   x_N(t+1)=f_N([x_j(t)]_{j\in N_N})
  \end{array}\right]\in \mathcal{D}^N
 \end{equation}
}

\textcolor{blue}{Each logical function $f_i$ has $l_i$ possibilities} and is chosen from the finite set of Boolean functions $\mathbf{F}_i=$ \textcolor{blue}{$\{f_i^1, f_i^2, \dots, f_i^{l_i}\}$} (hence, \textcolor{blue}{$|\mathbf{F}_i|=l_i$)} that the node is associated with.

Each function \textcolor{blue}{$f_i^{k}\in\mathbf{F}_i$, $k \in [1, l_i]$} is chosen with probability $Pr[f_i=f_i^k]=\textcolor{blue}{p_i^k}$ with $\sum_{k=1}^{l_i}\textcolor{blue}{p_i^k}=1$. \textcolor{blue}{In this article, we assume that the assignment of logical functions for each node $i$ is independent. Hence, }the probability of Boolean function selections, over $N$ nodes, is given by \textcolor{blue}{the product $p_1^{\mu_1} \cdot p_2^{\mu_2} \dots\cdot p_N^{\mu_N}$, where $\mu_i \in [1, l_i]$}. Different \textcolor{blue}{$f_i^k$} selections lead to different PBN realizations, 
which occur under different probabilities, 
resulting in stochastic state evolution of the network. Consequently, the possible realizations of a PBN are defined as $\mathcal{R} =\prod_{i=1}^N \textcolor{blue}{l_i}$.

Thus, the probability $P_{\mathbf{\mathcal{X}_t},\textcolor{blue}{\mathbf{\mathcal{X}_{t+1}}}}$ of transitioning from state $\mathbf{\mathcal{X}_t} \rightarrow \textcolor{blue}{\mathbf{\mathcal{X}_{t+1}}}$ at the next time step is: $P[x_1(t+1)\textcolor{blue}{,}  x_2(t+1),\dots, x_N(t+1) | x_1(t),x_2(t),\dots, x_N(t)] = P_{\mathbf{\mathcal{X}_t},\textcolor{blue}{\mathbf{\mathcal{X}_{t+1}}}}$ where $P[x_1, x_2, \dots, x_n | y_1, y_2, \dots, y_n]$ \textcolor{blue}{denotes} the joint probability of $x_1, x_2, \dots, x_n$ conditioned to $y_1, y_2, \dots, y_n$.
We can now construct the transition probability matrix $\mathcal{P}$ comprised of $2^N \times 2^N$ entries where entry \textcolor{blue}{$\mathcal{P}_{m,n}$} indicates the probability of transitioning from current state $m$ to possible next state $n$.

\subsection{\textcolor{blue}{Control problem formulation}}
\label{sec:control-problem}

In the context of PBNs, and consequently GRNs, control takes the form of discovering policies, or series of interventions (perturbations) to the state of a node (gene), aiming to drive the network from its current state to a desirable state, where the network exhibits desirable biological properties. 

\textbf{Definition 1. } \textit{Consider a PBN at state $\mathbf{\mathcal{X}_t}$. Then, define intervention} $\textrm{I}(\mathbf{\mathcal{X}_t}, \textcolor{blue}{i_t})$, $0\leq i \leq N$, \textit{as the process of flipping the binary value $x_i(t)$ associated with node $i$, at time $t$. \\ 
$i_{\textcolor{blue}{t}}=0$ denotes no intervention \textcolor{blue}{at time $t$}.}

Since an intervention strategy in gene therapies should be the least intrusive to the GRN,  only a single $\textrm{I}(\mathbf{\mathcal{X}_t}, i\textcolor{blue}{_t})$ is allowed at each time step $t$. This means an intervention is followed by a natural network evolution step according to the PBN internal transition dynamics\footnote{We stress that \textcolor{blue}{in this article} the probability distribution of successor states, from each state, is unknown when learning the control policy.}. Hence, we resist operating in a more aggressive intervention mode although it is favourable from a computational viewpoint.

\textbf{Definition 2. } \textit{Consider} $\textrm{I}(\mathbf{\mathcal{X}_t}, i_{\textcolor{blue}{t}})$, \textit{and intervention horizon} $H \in \mathbf{Z^+}$. \textit{Define} \textcolor{blue}{$\mathbf{S}=\{\textrm{I}(\mathcal{X}_1, i_{\textcolor{blue}{1}}), \textrm{I}(\mathcal{X}_2, i_{\textcolor{blue}{2}}), \dots \textrm{I}(\mathcal{X}_{h\leq H}, i_{\textcolor{blue}{h}})\}$, where $0\leq i_{\textcolor{blue}{t}} \leq N$, and $h \leq H \in \mathbf{Z^+}$. Again, $i_{\textcolor{blue}{t}}=0$ denotes no intervention, {\textcolor{blue}{at time $t$}}.}

The objective is to obtain the sequence of interventions $\mathbf{S}$ that directs the network from the current state - sampled from a uniform distribution of all states $p\textcolor{blue}{(\mathcal{X}_0)}$ - to a desirable state. 

\textcolor{blue}{When control input nodes are not known,} $\mathbf{S}$ is formed by perturbations effected on any node, hence there are $N+1$ actions for the RL agent at each state. The objective is to determine the sequence $\mathbf{S}$ within a finite number of interventions, the horizon $H$, assuming that the MDP is ergodic. The experiments reported in the next section show this not to be a  restricting assumption. 

\textcolor{blue}{When control input nodes are known, or can be computed,} $\mathbf{S}$ is formed by perturbations effected only on the control nodes. The objective here is to determine the sequence $\mathbf{S}$ required to increase the steady-state probability mass of desirable network states. In the study of control of the melanoma PBN \cite{par-2006-pbn7,sirin-2013-BatchRL} these are the states where the gene \textit{WNT5A}, \textcolor{blue}{which is central to the induction of an invasive phenotype in melanoma cells,} is OFF. Perturbations are only allowed on the \textit{pirin} gene, as in \cite{par-2006-pbn7,sirin-2013-BatchRL}. 
Since the target domain does not assume knowledge of attractors, this allows addressing larger networks (cf. we demonstrate control of the Melanoma PBN $N$=200).

\subsection{Deep Reinforcement Learning}
\label{sec:drl}

The central task of Reinforcement Learning \cite{sutton-2018-reinforcement} is to solve sequential decision problems by optimising a cumulative future reward. This can be achieved by learning estimates for the optimal value of each action which is typically defined as the sum of future rewards when taking that action and following the optimal policy afterwards. 

The strategy that determines which action to take is called a \textit{policy}.
Hence, an \textit{optimal policy} results from selecting the actions that maximize the future cumulative reward.

Q-Learning \cite{watkins-1992-q} maintains an estimate of the optimal value function $Q: \mathcal{S} \times \mathcal{A} \rightarrow \mathbb{R}$ which is updated towards the target:
\begin{equation}
\label{eq:classicQL}
	\begin{split}
		Q(s_t, a_t) &\leftarrow Q(s_t, a_t) + \\
		&+ \alpha[r_{t+1} + \gamma {max}_{a^{'}} Q(s_{t+1}, a^{'})- Q(s_t, a_t)]
	\end{split}
\end{equation}
where $Q(s_t, a_t)$ is the expected reward of taking action $a_t$ at state $s_t$ at time step $t$, 
$r_{t+1}$ is the reward received at the next time step after taking action $a_t$, $\gamma$ is the discount factor that trades off the importance of immediate and later rewards, 
$r_{t+1} + \gamma {max}_{a^{'}}Q(s_{t+1}, a^{'})$ is the TD\footnote{Temporal-Difference (TD) learning refers to algorithms where estimates at the current time step are based on those of the next time step \cite{sutton-2018-reinforcement}.}-Target, 
$r_{t+1} + \gamma {max}_{a^{'}}Q(s_{t+1}, a^{'}) - Q(s_t, a_t)$ \textcolor{blue}{is the TD-Error} \textcolor{blue}{$\delta$}, and $0 < \alpha \leq 1$ is a constant that determines how fast the agent forgets past experiences.
Contextually, given state $s_t$, Eq. (\ref{eq:classicQL}) improves the estimate of the value associated with action $a_t$. Note Eq. (\ref{eq:classicQL}) does not use any probabilities defined by the MDP.

The true value of each state-action pair, $Q^{*}(s,a)$ can be approximated iteratively by selecting actions at each time step.
In this work we select actions following the $\varepsilon$-greedy \textit{policy} algorithm, where $\varepsilon$ is the probability of randomly selecting an action $a$ at each time step, and $(1-\varepsilon)$ of greedily performing the action with the maximum expected reward determined by ${max}_aQ(s_t, a)$.
The value of $\varepsilon$ starts from $1$ and decreases towards a constant value ${min}_\varepsilon$ at every time step.
After an action is selected $Q(s_t, a_t)$ is updated according to Eq. (\ref{eq:classicQL}).
Note that $Q$ has been shown to converge to $Q^{*}$ \cite{sutton-2018-reinforcement,watkins-1992-q}.

 The objective is to determine a policy $\pi : \mathcal{S} \rightarrow \Delta_{\mathcal{A}}$ that maximises the expected return $\mathcal{J}$, for each initial state-action pair $(s,a) \in \mathcal{S} \times \mathcal{A}$, by direct interaction with the environment: 
 \begin{equation}
 \label{eq:policy}
  \mathcal{J} = \mathbb{E}_\pi[r(s, a)] = \mathbb{E}[\sum_{t=0}^{H-1}\gamma^tr(s_t, a_t) | \mathcal{P}, \pi]
 \end{equation} where $H$ is the intervention horizon.
  A range of methods have been studied for this problem \cite{Tsitsiklis-1996-neuroDP}, \cite{sutton-2018-reinforcement}. Here, we \textcolor{blue}{appeal to} a model-free RL method which draws upon Double Deep Q-Learning, with experience replay.  

\textbf{Double Deep Q Network (DDQN).}
Deep Reinforcement Learning combines classical reinforcement learning with neural network function approximation. More specifically, Double Q-Learning \cite{van-hasselt-2010-DoubleQLearning} is combined with Deep Q-Learning \cite{mnih-DeepQ-Learning} to address the issue of overestimating $Q$ values. 
A parametric form of the state-action value function of  Eq. (\ref{eq:classicQL}), \textcolor{blue}{$Q(s,a; \theta)$}, with parameters $\theta$ is often represented using neural networks: 
\begin{equation}
    \label{eq:parametric}
    \begin{split}
        \theta_{t+1} = \theta_t + \textcolor{blue}{\alpha}[\textcolor{blue}{r}_{t+1} \textcolor{blue}{+} \gamma{max}_{a^{'}}Q(s_{\textcolor{blue}{t+1}}, a\textcolor{blue}{^{'}}; \theta_t) - \\ 
        - Q(s_t, a_t; \theta_t)]\nabla_{\theta_t}Q(s_t, a_t; \theta_t)
    \end{split}
\end{equation}
A Deep Q Network (DQN) is used here for this purpose, which is a multi-layer neural network with inputs being the current state observation of the environment and outputs a vector of the expected action values for that state $Q(s, \cdot; \theta)$.
Training involves direct interaction with the (PBN) environment.
The goal of the DQN is to iteratively update $\theta$ in order to approximate $Q^{*}(s,a;\theta)$.
The DQN is trained by minimizing a (\textit{different}) sequence of loss functions at each iteration:
\begin{equation}
\label{eq:loss}
	L(\theta_t) = (\textcolor{blue}{r}_{t+1} + \gamma {max}_{a^{'}}Q(s_{t+1}, a^{'}; \theta^{-}_t) - Q(s_t, a_t; \theta_t))^2
\end{equation}

where $\theta_t$ denotes the parameters of \textcolor{blue}{$Q$ of Eq. (\ref{eq:classicQL})} and $\theta^{-}_t$ is a periodic copy of $\theta_t$. 
It is worth noting here that following the suggestion in  \cite{mnih-DeepQ-Learning} a separate network is used to determine the TD-Target. This \textit{target} DQN is initialised with the same parameters as the main DQN (the so-called \textit{policy} DQN), but has its parameters updated every $k$ iterations. 
\textcolor{blue}{That is, the expected $Q$ values of the \textit{target} DQN are fixed and every $k$ iterations the parameters of the \textit{policy} DQN are copied to the \textit{target} DQN.}

By differentiating Eq. (\ref{eq:loss}) we obtain:
\begin{equation}
\label{eq:lossgrad}
	\begin{split}
		\nabla_{\theta_t}L(\theta_t) &= (r_{t+1}+\gamma{max}_{a^{'}} Q(s_{t+1}, a^{'};\theta_t^{-}) \\
		&- Q(s_t, a_t; \theta_t))\nabla_{\theta_t}Q(s_t, a_t;\theta_t)
	\end{split}
\end{equation}
which can be used to update the DQN parameters using stochastic gradient descent.
In the context of Deep Reinforcement Learning \cite{van-hasselt-2010-DoubleQLearning} two networks corresponding to the two Q functions are used (as in \cite{van-2016-deep}) and Eq. (\ref{eq:lossgrad}) becomes:
\begin{equation}
\label{eq:lossDRL}
	\begin{split}
		L(\theta_t) &= (\textcolor{blue}{r}_{t+1} + \gamma Q(s_{t+1}, {argmax}_{a^{'}}Q(s_{t+1},\textcolor{blue}{ a^{'};\theta_t});\theta_t^{-}) \\
		&- Q(s_t, a_t; \theta_t))^2
	\end{split}
\end{equation}
The update process remains the same where the parameters $\theta_t$ are copied to $\theta_t^{-}$ periodically, every $k$ iterations.

\textbf{Prioritized Experience Replay (PER).} 
During the training (cf Section \ref{sec:ddqn-control-pbn}) the agent interacts with the environment by observing state $s_t$ and performing action $a_t$, either randomly with probability $\varepsilon$ or greedily with probability $(1-\varepsilon)$ by selecting the action with the highest $Q$ value, then the environment transitions to state $s_{t+1}$ and the agent receives reward $\textcolor{blue}{r}_{t+1}$. Thus, the agent experiences a sequence of transition tuples $(s_t, a_t, r_{t+1}, s_{t+1})$ which are stored in a replay buffer $\mathcal{B}$. The loss function of Eq. (\ref{eq:loss}) is obtained as the expectation over a batch of such tuples capturing the experiences of the agent. Every time step $t$ a batch of experiences is sampled from \textcolor{blue}{$\mathcal{B}$} and used to update the Double DQN parameters using Prioritized Experience Replay (PER) \cite{schaul-2015-PER}.

Remember our approach does not use the Probability Transition Matrix, i.e., it is {\it model-free}, hence the agent learns the control policy by sampling the environment. However, training a network from consecutive samples directly obtained from the environment is inefficient because of the strong correlations in this data, which can cause a high variance in the network parameter updates \cite{mnih-DeepQ-Learning}.

Also, rare experiences can be forgotten rapidly because of the way $Q$ values are updated. Such experiences are useful because if the sampled experiences are dominated by those frequently occurring, the Double DQN can become biased towards selecting actions useful only for those experiences. PER breaks this correlation and avoids large oscillations of the network parameters leading to more stable learning, while allowing experiences to be seen more often, hence, also avoiding rare experiences from being forgotten fast \cite{schaul-2015-PER}.

In our implementation we use \emph{proportional} prioritization compared to rank-based prioritization, as experimentally it yielded slightly better results. The probability $P(i)$ of a tuple $i$ being sampled is given by: 
\begin{equation}
    P(i) = \frac{p^\omega_i}{\sum_{z\in\mathcal{B}} p^{\omega}_z}
\end{equation}
where \textcolor{blue}{$p_i$ is the priority value of the $i$-th tuple, given by $p_i = |\delta_i| + c$, where $|\delta_i|$ is the absolute TD-error associated with the $i$-th tuple, $c$ is a small constant to prevent experiences with zero TD-Error from never being replayed}, and exponent $\omega$ determines the magnitude of prioritization, \textcolor{blue}{with $\omega = 0$ corresponding to the uniform case}.

However, prioritizing replay creates a bias in learning towards samples with high TD-Error as they are sampled more often. To avoid this, importance weights \cite{schaul-2015-PER} are attached to each tuple, given by:
\begin{equation}
\label{eq:imp-weights}
 w_i=(\frac{1}{|\mathcal{B}|\cdot P(i)})^\beta
\end{equation}
where $|\mathcal{B}|$ is the size of the buffer, and $\beta$ is a hyperparameter used to anneal the amount of importance sampling over training episodes.

\section{\textcolor{blue}{Deep RL to control large PBNs}}
\label{sec:ddqn-control-pbn}

To address the control problem, a Deep Neural Network was constructed with an input layer of size $N$, for a PBN with $N$ nodes, two hidden layers with $M$\textcolor{blue}{\footnote{The number of rectifier units $M$ depends on $N$; through experimentation we found that $M=64$ worked well in the PBNs under study (Section \ref{sec:results-exp}).}} rectifier units each, and a linear output unit of size $N+1$ where $N$ corresponds to the expected Q values of the $N$ possible perturbations at each time step, while the extra unit corresponds to taking no action, i.e., no perturbation.

The behavior policy \textcolor{blue}{applied} during training was $\varepsilon$-greedy \cite{mnih-DeepQ-Learning} \textcolor{blue}{and our parameters are largely in line with the literature.} $\varepsilon$ \textcolor{blue}{was set to start} at $1$ and decay to \textcolor{blue}{$0.05$ over a given fraction of the training steps}, where it stays fixed. \textcolor{blue}{In practice this fraction was set to $0.75$ when the target domain was an attractor, and $0.1$ and $0.5$ when the target was a pre-assigned subset of the state space (depending on the PBN).}
The selected discount factor $\gamma$ was set to $0.99$ to weigh future rewards strongly, as the agent may direct the network to the desired state after a large number of steps.

For Prioritized Experience Replay we set $\omega=0.6$ and linearly anneal $\beta$ from $0.4$ to $1$ after $75\%$ of training as done in \cite{schaul-2015-PER}.
Experimentally these hyperparameter choices yielded the best results. Intuitively, we want to prioritize aggressively, but not dominate the sampled experiences with large TD-Error experiences.
The size of the replay buffer selected \textcolor{blue}{varies depending on the experiment, see Section \ref{sec:results-exp}}.

During training the algorithm used for optimizing the network parameters is \textcolor{blue}{Adam}, with $\alpha$ = \textcolor{blue}{0.0001}.
Huber loss is used to minimize the network's  TD-Error.
The reason behind this choice is to avoid exploding gradients by error clipping \cite{faryabi-2007-RL-GRNs}.
We also sample a batch of \textcolor{blue}{$256$ or $128$} experiences from the replay memory every time we update the network parameters.
\textcolor{blue}{Please see the Supplementary Material for further details on hyperparams.}

\textbf{Reward Scheme \textcolor{blue}{\& Episode Design}} Measured design of the reward assignment is required to approximate $Q(s,a; \theta)$ during training and guide the agent towards the goal, i.e., derive the shortest sequence of perturbations that direct to the target domain.
Consider $\mathcal{Y}$ to be the set of desirable states, $s_t=\mathbf{\mathcal{X}_t}$ and $a_t=i$.
We define the following
reward function:
\begin{equation}
\label{eq:reward}
r(s_t,a_t):=\begin{cases} 
      r > 2 & \textrm{if } s_{t+1}\in\mathcal{Y}  \\
      -2 & \textrm{if } s_{t+1} \textrm{ is in a undesirable attractor} \\
      -1 & \textrm{if } s_{t+1} \textrm{ is any other state}
   \end{cases}
\end{equation}

\textcolor{blue}{Moreover, there is an additional cost to the agent taking an action by returning $r(s_t,a_t)-1$ if $a_t\neq 0$}. The action cost encourages minimum interventions. This scheme leads to successful control, as shown in the experiments (Section \ref{sec:results-exp}). 

\textcolor{blue}{When the target domain is an attractor, an episode ends either when the desired attractor $\mathcal{Y}$ is reached or the horizon $H$ has been reached.}
The PBN dynamics entail that the network gravitates towards some attractor state, which may not be the desirable one. In such cases, the policy gets a reward of -2 so as to learn to avoid transitioning to non-desirable attractor states. Finally, to compensate for the fact the desirable state will rarely be achieved initially, especially in larger state spaces, the policy gets a reward $r(s_t, a_t)>2$ whenever it does achieve it - in practice, this was set to $5$. This encourages experience tuples with high priority to be sampled more often due to the PER mechanism.

Note that in cases where only the desirable attractor is known the scheme can be readily adapted so that the policy gets a reward of -1 for all states except for those in the desirable attractor. It may be instructive to read Eq. (\ref{eq:reward}) in terms of the policy $\pi$ maximising the expected return $\mathcal{J}$ of Eq. (\ref{eq:policy}) discussed earlier. 

Following Eq. (\ref{eq:reward}), maximising the performance objective $\mathcal{J}$  of Eq. (\ref{eq:policy}) corresponds to the process of finding a sequence of perturbations that drive the PBN to the desirable states $\mathcal{Y}$ and achieve that with the least possible perturbations.

\textcolor{blue}{When the target domain is a pre-assigned subset of the state space, the reward scheme and episode design are slightly adjusted. The positive reward for being in a favourable state is retained, but in practice set to a larger value, $10$. Additionally, there is no notion of an ``undesirable attractor'' now. Instead, there is a larger cost for being in an undesirable state, namely the inverse of the one for being in a desirable state, i.e., $-10$. Finally, the action cost is retained. Similar to typical RL continuous control environments, an episode is only terminated in this case when the horizon $H$ has been reached. 
}

\textbf{Training. } Following standard practice, we train the network for a number of episodes--or time steps--and update the second estimator (\textit{target} DQN) periodically.

\begin{figure}[h]
	\centering
\scalebox{0.75}{\includegraphics[width = 0.7\linewidth]{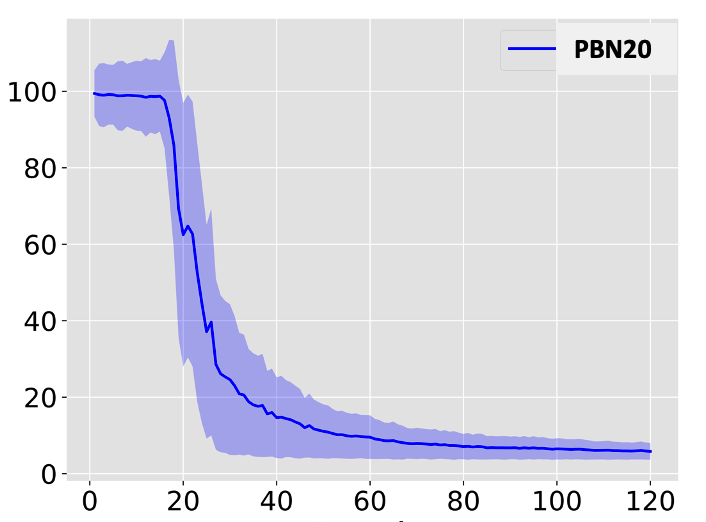}}
	\caption{Avg number of perturbations (y-axis) per epoch (x-axis) decreases as agent learns control policy during training (shown here for PBN $N$=20)}
	\label{fig:avg-int-per-epoch-PBN20}
\end{figure}

\textcolor{blue}{When the target domain is an attractor, we train for up to 670,000 episodes in total depending on the network, updating the \textit{target} DQN every 5,000 episodes, which we call an ``epoch''.} During training we evaluate the performance of the learning agent by plotting the average number of perturbations per epoch, and the average reward per epoch. Fig. \ref{fig:avg-int-per-epoch-PBN20} shows the average number of perturbations decrease as training evolves, hence the agent learns the policy for the target network.

It may be worth noting the sharp drop in the average number of perturbations around the 18th epoch, i.e., after the first 90,000 episodes. \textcolor{blue}{We return to this point when we discuss stabilization of the $N$=20 PBN (cf. Section \ref{sec:results-exp})}.


\textcolor{blue}{When the target domain is a subset of the state space, we opt to train for a number of time steps rather than a number of episodes, as the episode itself has lost the semantics attached to it. We train for 150,000 time steps and update the \textit{target} DQN every 10,000 time steps with the exception of the $N$=70 PBN where we do so every 1,000 time steps. 
}

\textbf{Successful control.} 
After training is complete, \textcolor{blue}{when the target is an attractor,} we evaluate our control method by initializing the PBN in question from every possible state and attempting to direct it to the desirable state multiple times. Each attempt terminates when (i) the desirable attractor (fixed point or cyclic) is reached, or (ii) the agent exhausts the number of perturbations available to it, i.e., exceeds the horizon $H$. 
Whenever the attractor is reached, this counts as a successful control attempt.

\textcolor{blue}{When the target is a subsrt of the state space, we evaluate our method by computing the steady state distribution histogram of the network averaging across 300 distinct runs of 4,000 time steps each.}

The set up and training of the DDQN with PER is given in Algorithm \ref{alg:ddqnper}, which is what has been implemented to control the PBNs discussed in the following section. 

\textcolor{blue}{Time complexity has been challenging in existing works. The time complexity of Algorithm \ref{alg:ddqnper} is best approximated by: 
\begin{equation}
   \mathcal{O}(t ^. \mathcal{B} ^. |\theta|) 
\end{equation}
where $t$ is the number of time steps, $\mathcal{B}$ is the batch size (samples fed into the neural network (of the DQN) for gradient descent), and $|\theta|$ is the number of parameters in the neural network.}

\textcolor{blue}{It can be seen that our approach is {\it linearly} dependent on the time steps needed in training, for learning the optimal policy in reaching the target domain (selecting actions with the maximum reward). While this grows with the size of the network, the time complexity of the approach is a significant advance to existing approaches which are exponential or polynomial. This is in fact illustrated in the following section.}
    


\begin{algorithm*}
    \caption{DDQN with PER Training Algorithm}\label{alg:ddqnper}
    \textbf{Input}: $\gamma$, $\min_\epsilon$, $|\mathcal{B}|$, $\beta$, $\omega$, $\alpha$, $N$, $N_{episodes}$, $N_{epochs}$, \texttt{horizon}, \texttt{batchSize}, c, \texttt{updateInterval}
    \newline\textbf{Output}: $\theta^{*}$
    \begin{algorithmic}
        \State $\theta \leftarrow \texttt{rand}([0,1]),\,\theta^{-} \leftarrow \theta$\Comment{Initialize network weights}
        \State $\mathcal{B} \leftarrow \varnothing,\,\texttt{inc}_\beta \leftarrow \frac{\beta}{0.75 \times N_{episodes} \times N_{epochs}}, \max_p \leftarrow 1$\Comment{Initialize PER}
        \State $\epsilon \leftarrow 1,\,\texttt{dec}_\epsilon \leftarrow \frac{1 - \min_\epsilon}{N_{episodes} \times N_{epochs}}$\Comment{Initialize $\epsilon$-greedy}
        \State $\texttt{trainCount} \leftarrow 0$
        \For{\texttt{epoch}$\,\in[0,N_{epochs}]$}
            \For{\texttt{episode}$\,\in[0,N_{episodes}]$}
                \State $t \leftarrow 0, s_t \leftarrow \varnothing$
                \State $\mathbf{\mathcal{X}_t} \leftarrow \texttt{rand}(\mathcal{D}^N)$\Comment{Initialize PBN to a random state}
                \While{$s_t \notin \mathcal{Y}\,\land \,t \neq \texttt{horizon}$}
                    \State $s_t \leftarrow \texttt{read}(\mathbf{\mathcal{X}_t}),\,a_t \leftarrow \epsilon\texttt{-greedy}(\epsilon, s_t)$
                    \State $s_{t+1},\, r(s_t, a_t) \leftarrow \texttt{apply}(\texttt{action})$\Comment{Apply the chosen action to the environment}
                    \State $\texttt{saveToReplayBuffer}(\mathcal{B},\,(s_t, a_t, r(s_t, a_t), s_{t+1},\,\max_p))$
                    \If{$|\mathcal{B}| \geq \texttt{batchSize}$}
                        \State $\text{Sample }(\mathbf{T}=\{\mathbf{S}, \mathbf{A}, \mathbf{R}, \mathbf{S}^\prime\},\,\mathbf{W})\text{ where }\forall i\in\mathbf{T},\,P(i)=\frac{p_i^\omega}{\sum_{z\in\mathcal{B}}p_z^\omega},\forall w_i \in \mathbf{W},\,w_i=(|\mathcal{B}|\cdot P(i))^{-\beta}$
                        \State $L(\theta) \leftarrow (\mathbf{R} + \gamma \max_{a^\prime}Q(\mathbf{S}^\prime, a^\prime; \theta^{-}) - Q(\mathbf{S}, \mathbf{A}; \theta)) \cdot \mathbf{W}$
                        \State $\theta \leftarrow \theta - \alpha \nabla_\theta L(\theta)$\Comment{Gradient Descent}
                        \State $\forall i \in \mathbf{T}$ update priorities with $p^\prime_i \leftarrow L(\theta) \times c$, $\max_p \leftarrow \max(p_i \forall i \in \mathcal{B})$\Comment{Update PER priorities}
                        \State $\texttt{trainCount} \leftarrow \texttt{trainCount} + 1$
                        \If{$\texttt{trainCount} \mod \texttt{updateInterval} = 0$}\Comment{Update second DQN}
                            \State $\theta^{-} \leftarrow \theta$
                        \EndIf
                    \EndIf
                    \State $t \leftarrow t + 1$
                \EndWhile
                \State $\beta \leftarrow \min(\beta + \texttt{inc}_\beta, 1),\,\epsilon \leftarrow \max(\epsilon - \texttt{dec}_\epsilon, \min_\epsilon)$\Comment{Update annealed parameters}
            \EndFor
        \EndFor
    \end{algorithmic}
\end{algorithm*}

\section{Experiments \& Results}
\label{sec:results-exp}

We report on experiments of applying the model-free deep RL (DDQN with PER) \textcolor{blue}{control} method \textcolor{blue}{outlined} in this paper on a number of networks \textcolor{blue}{of varying sizes (e.g., 7, 20, 28, 70, 200 nodes) and cover both the case where the goal is to reach a specific attractor and where the goal is to shift the probability mass favourably}. More specifically, we report on applying it to two \textcolor{blue}{synthetic} PBNs of $N$=10 and $N$=20, as well as four PBNs generated directly from gene data \textcolor{blue}{studied in the literature}, with $N$=7, $N$=28, $N$=70 \textcolor{blue}{ and $N$=200}.

 \textcolor{blue}{These experiments were run on standard issue hardware and took anywhere from 20 minutes to 6 hours depending on the training length.}

\subsection{\textcolor{blue}{Attractor states as the target}}
\label{sec:attractor-control}

We start with results on PBNs where the agent was allowed to perform perturbations on any node of the network. 

\textbf{PBN $N$=10.} The connectivity of the PBN with 10 nodes, is given as follows: ${inp}^1=[1,10], {inp}^2=[3, 8], {inp}^3=[8, 10], {inp}^4=[7, 8], {inp}^5 = [9, 6], {inp}^6 = [8, 2], {inp}^7 = [10, 4], {inp}^8 = [5, 9], {inp}^9 = [10, 9], {inp}^{10} = [4, 7]$
where ${inp}^i: N \rightarrow [1,N]$ denotes the set of nodes that provide input to node $i$.

We compute the attractors in a standard way, e.g., using the  NetworkX\textcolor{blue}{\cite{hagberg2020networkx} library}, and find this PBN has 3 attractors, namely the network states $A_1 = \{0000000000\}$, $A_2 = \{10000000000\}$ and a cyclic attractor  $A_3$ comprising 143 states $A_3 = \{1111001111, 1111001110, \dots, 1010010110, 1000101110\}$;
this can be easily reproduced by rolling out the PBN with the connectivity given earlier and the dynamics shown in Table \ref{table:pbfuncsn10}.
\begin{table}[h]
\begin{center}
	\scalebox{0.9}{\begin{tabular}{|c|c c c|c|c c c|}
		\hline
		$\mathbf{F}_i$ & \textbf{OR} & \textbf{AND} & \textbf{XOR} & $F^i$ & \textbf{OR} & \textbf{AND} & \textbf{XOR} \\
		\hline
		$\mathbf{F}_1$ & $1$ & - & - & $\mathbf{F}_6$ & $0.82$ & $0.15$ & $0.03$ \\
		$\mathbf{F}_2$ & $0.5$ & $0.25$ & $0.25$ & $\mathbf{F}_7$ & $0.48$ & $0.52$ & - \\
		$\mathbf{F}_3$ & $0.71$ & $0.29$ & - & $\mathbf{F}_8$ & $0.28$ & $0.45$ & $0.27$ \\
		$\mathbf{F}_4$ & $0.52$ & $0.48$ & - & $\mathbf{F}_9$ & $1$ & - & - \\
		$\mathbf{F}_5$ & $0.36$ & $0.05$ & $0.59$ & $\mathbf{F}_{10}$ & $0.99$ & $0.01$ & - \\
		\hline
		 & \textcolor{blue}{$p_i^1$} & \textcolor{blue}{$p_i^2$} & \textcolor{blue}{$p_i^3$} &  & \textcolor{blue}{$p_i^1$} & \textcolor{blue}{$p_i^2$} & \textcolor{blue}{$p_i^3$} \\
		 \hline
	\end{tabular}}
	\caption{\label{table:pbfuncsn10}Dynamics of the PBN $N$=10; each function $\mathbf{F}_i$, $i=1..10$, with the probability \textcolor{blue}{$p_i^{1..3}$} of it being assigned to node $i$ is shown.}
\end{center}
\end{table}

In order to select a target attractor, we allowed the PBN to \textit{naturally} evolve assuming a uniform starting distribution for all states. We noted that $A_1$ only occurs with probability $P=0.0097$, $A_2$ with probability $P=0.0107$ and $A_3$ with probability $P=0.9796$. This was to be expected as attractor $A_3$ comprises 144 network states.
In our experiments with this PBN we set our desired state to be attractor $A_1$ as its probability of \textit{naturally} occurring is the least.

In order to set the horizon $H$, we attempt to control the PBN starting from every possible state multiple times by applying random perturbations and note that the average interventions needed was $1\textcolor{blue}{,}387$.
We decided to set the horizon at $11$ perturbations - that is, approximately $0.8\%$ of the average random perturbations needed to land on $A_1$.
This PBN makes an interesting case for controllability as its possible network realizations, at each time step, are $N$=1,296, which makes state transitions highly non-deterministic.

For this experiment, with reference to \textcolor{blue}{Section \ref{sec:ddqn-control-pbn}}, we run 300,000 episodes for training, we set the reward $r$=5, the discount factor $\gamma$ to 0.99, parameter $c$ in PER to 500. The size of the input and output units have been set to match the size of the PBN and \textcolor{blue}{number of possible perturbations, i.e., 11}. The size of the memory buffer was set to $10,000$ and we sampled experiences in batches of $128$. Finally, we let the network train for $300,000$ episodes and update the
\textcolor{blue}{second estimator network} every $400$ episodes.

The agent achieves on average 99.8\% successful control on the horizon of 11 perturbations. This result indicates that the network fails to always control 0.02\% of the possible PBN states. As already mentioned in Section \ref{sec:ddqn-control-pbn} there is a tradeoff between the number of allowed perturbations and maximizing the probability of \textcolor{blue}{successful control}.  It is interesting to note that if we allow the same DDQN \textcolor{blue}{with PER} (trained on a maximum of 11 interventions) to perform 14 interventions during testing, it achieves 100\% successful control. 

The reason behind this result lies in the mechanics of Q-Learning. As explained in Section \ref{sec:prelimin}\textcolor{blue}{C}, Q-Learning aims to maximize the cumulative reward. Hence, even if the agent was not successful in finding a policy that always achieves 100\% control by at most 11 perturbations, it manages after the 11th perturbation to drive the PBN to a state that has previously been shown to evolve to states that are controllable. Thus, if the state naturally occurring after the 11th intervention has previously been controlled and the agent is allowed perturbation on that state, it will successfully drive it to the desired attractor.

\textcolor{blue}{The fact that 14 perturbations was the maximum number required to achieve control, with 100\% success rate, means that the DDQN drives the states that failed to be controlled within $11$ perturbations to} states that at the next time step will always naturally transition to states that can be controlled within a maximum of $3$ perturbations.

\begin{figure}[h]
	\centering
\scalebox{0.75}{\includegraphics[width = 0.7\linewidth]{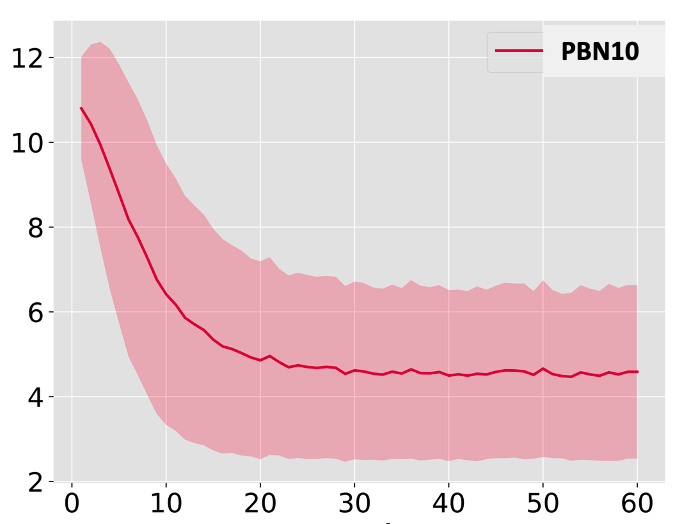}}
	\caption{The avg number of perturbations (y-axis) per epoch (x-axis) decrease as the agent determines the control policy (PBN $N$=10).}
	\label{fig:avg-interv-per-epoch-PBN10}
\end{figure}

\textcolor{blue}{Further, it can be seen in} Fig.\ref{fig:avg-interv-per-epoch-PBN10} that the average number of perturbations needed to control the PBN begin to decrease after training has begun - sharp decrease after the first 10 epochs in Fig.\ref{fig:avg-interv-per-epoch-PBN10}.  This would not be justifiable if the RL agent did not succeed in finding a control policy that drives the PBN to the desired attractor \textcolor{blue}{within the horizon of $11$ perturbations.}

\textbf{PBN $N$=20.}  The connectivity of the PBN with 20 nodes is as follows: ${inp}^1=[3,6],
{inp}^2=[7, 14], {inp}^3=[3, 5], {inp}^4=[7, 4], {inp}^5=[9, 6], {inp}^6=[3, 11],
{inp}^7 = [11, 3], {inp}^8 = [10, 9], {inp}^9 = [14, 7], {inp}^{10} = [8, 19],
{inp}^{11}=[8,6], {inp}^{12}=[9, 4], {inp}^{13}=[14, 16], {inp}^{14}=[14, 18], {inp}^{15} = [19, 15],
{inp}^{16} = [19, 2], {inp}^{17} = [18, 4], {inp}^{18} = [1, 20], {inp}^{19}=[2, 5], {inp}^{20}=[18, 20]$, where, \textcolor{blue}{again,} ${inp}^i$ denotes the set of nodes that provide input to node $i$. 

Note \textcolor{blue}{the model-free deep RL method we propose does} not require knowledge of the dynamics of the environment, in terms of the distribution of probabilities of successor states, from each state. However, we include here the Boolean function probabilities \textcolor{blue}{(see Table \ref{pbnfuncsn20})} and the PBN structure to aid with our experiments being reproduced.

\begin{table}[h]
\begin{center}
\scalebox{0.9}{\begin{tabular}{|c|c c c|c|c c c|}
		\hline
		$\mathbf{F}_i$ & \textbf{OR} & \textbf{AND} & \textbf{XOR} & $\mathbf{F}_i$ & \textbf{OR} & \textbf{AND} & \textbf{XOR} \\
		\hline
		$\mathbf{F}_1$  & $0.39$	 & $0.05$ & $0.57$ & $\mathbf{F}_{11}$ & - 	& $1$	& - \\
		$\mathbf{F}_2$  & $0.70$	 & -	  & $0.30$ & $\mathbf{F}_{12}$ & -	& $1$	& - \\
		$\mathbf{F}_3^3$  & $1$	 & -	  & -	   & $\mathbf{F}_{13}$ & $1$	& -	& - \\
		$\mathbf{F}_4$  & $0.18$	 & $0.82$ & -	   & $\mathbf{F}_{14}$ & $0.01$	& $0.98$& $0.01$ \\
		$\mathbf{F}_5$  & -	 & $0.11$ & $0.89$ & $\mathbf{F}_{15}$ & -	& -	& $1$ \\
		$\mathbf{F}_6$  & $1$	 & -	  & -	   & $\mathbf{F}_{16}$ & -	& $1$	& - \\
		$\mathbf{F}_7$  & $1$	 & -	  & -	   & $\mathbf{F}_{17}$ & $1$	& -	& - \\
		$\mathbf{F}_8$  & -	 & $0.44$ & $0.56$ & $\mathbf{F}_{18}$ & -	& $1$ 	& - \\
		$\mathbf{F}_9$  & -	 & -	  & $1$	   & $\mathbf{F}_{19}$ & -	& -	& $1$ \\
		$\mathbf{F}_{10}$& $0.82$ & $0.09$ & $0.09$ & $\mathbf{F}_{20}$ & $1$	& -	& - \\
		\hline
		 & \textcolor{blue}{$p_i^1$} & \textcolor{blue}{$p_i^2$} & \textcolor{blue}{$p_i^3$} &  & \textcolor{blue}{$p_i^1$} & \textcolor{blue}{$p_i^2$} & \textcolor{blue}{$p_i^3$} \\
		 \hline
	\end{tabular}}
	\caption{\label{pbnfuncsn20} Dynamics of the PBN with $N$=20 
	}
\end{center}	
\end{table}

Again, in order to select \textcolor{blue}{the specific desirable} attractor, we allowed the PBN to \textit{naturally} evolve assuming a uniform starting distribution of all states. The attractor $A=\{0,0,0,0,0,0,0,0,0,0,0,0,0,0,0,0,0,0,0,0\}$ was shown to occur the least. This was set as the desirable attractor so as to address the most difficult case for control. 

We attempted to control the PBN starting from every possible state multiple times by random perturbations and noted that the average perturbations needed was 6,511. The maximum number of allowed perturbations was set at approximately $1.5\%$ of the average random perturbations, i.e., \textcolor{blue}{a horizon $H=100$ perturbations}. 

Regarding the parameters of the DDQN with PER discussed in the previous section \textcolor{blue}{(Section \ref{sec:ddqn-control-pbn})}, for this experiment, the discount factor $\gamma$ was set to $0.99$, $c$ to 5,000 and the reward of Eq. (\ref{eq:reward}) was set to $r$=20. The size of the memory buffer was 500,000. Finally, we let the network train for 670,000 episodes and update the second estimator every 5,000 episodes.

The agent achieves $100\%$ control on the horizon of $100$ interventions.
Further, it is worth noting that the success rate only drops by a small fraction to just over $99\%$ if we limit the agent to 15 perturbations only. This explains the sharp drop on the (avg) number of perturbations seen earlier in Fig. \ref{fig:avg-int-per-epoch-PBN20}.

\textbf{Melanoma PBN $N$=7. } 
The proposed method was also applied to PBNs of real GRNs. We report here results on stabilization of a 7 node PBN, obtained from the gene expression data on metastatic melanoma found in \cite{bittner-2000-melanoma}, to a specific attractor. \textcolor{blue}{To allow for direct comparative study, the PBN includes the same 7 genes - {\it pirin, WNT5A, S100P, RET1, MART1, HADHB, and STC2} (in that order), which are also studied in existing literature  \cite{par-2006-pbn7,sirin-2013-BatchRL}}. 

The attractors in this melanoma PBN $N$=7 are $A_1 = 1001001$, $A_2 = 0110110$ and $A_3 = 0101111$. Drawing on domain knowledge that a deactivated WNT5A gene can reduce metastasis on the Melanoma GRN \cite{datta-2003-control}, the desired state in this experiment was set to be the attractor where WNT5A is $OFF$, i.e., is set to $0$, namely: $\textcolor{blue}{A_1 =}  1\mathbf{0}01111$.

\begin{figure}[h]
	\centering
\scalebox{0.65}{\includegraphics[width = 0.7\linewidth]{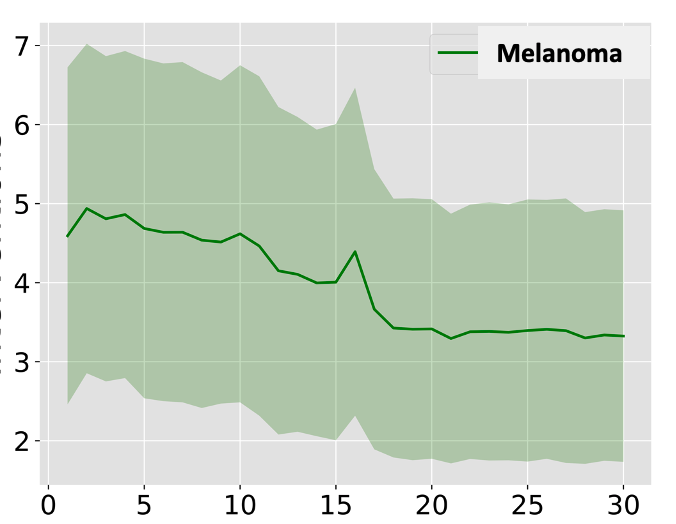}}
	\caption{The avg number of perturbations (y-axis) per epoch (x-axis) required to  control the Melanoma PBN $N$=7}
	\label{fig:avg-interv-per-epoch-PBN7}
\end{figure}

For this experiment, we run 150,000 episodes for training, we set the reward $r$=5, $\gamma$ to $0.9$, $c$ in PER to 500, and allowed up to 7 perturbations. The agent develops a control policy, see Fig. \ref{fig:avg-interv-per-epoch-PBN7}, and achieves $99.72\%$ control on the horizon of $H$=7 \textcolor{blue}{perturbations}. It achieves $100\%$ control if allowed up to $10$ \textcolor{blue}{perturbations}. More details on this experiment found in \cite{CN2020-drl-pbn}. 

In terms of comparative study, the proposed method produces a success rate of $99.72$\% compared to $96$\% in \cite{par-2006-pbn7} and $98$\% in \cite{sirin-2013-BatchRL} who incidentally employ a batch-type of Reinforcement Learning. We note the favourable comparison despite the fact the action space is reduced to perturbations on one node only (namely {\it pirin})  in both works \cite{par-2006-pbn7}, \cite{sirin-2013-BatchRL}  while we considered perturbations on any node here.

\subsection{\textcolor{blue}{Pre-assigned subset of the state space as the target}}
\label{sec:ssd-control}

 Next, we demonstrate the DDQN with PER control method \textcolor{blue}{in the case where the target is a pre-assigned subset of the state space. In addition, in this section} perturbations can only be performed by the agent on a subset of the nodes, the so-called {\it control} or {\it input} nodes. \textcolor{blue}{The term Probabilistic Boolean Control Networks (PBCNs) is often used in the literature.}
 
 We have tested the method on PBNs of varying dimension, i.e., $N$=7, $N$=28, $N$=70 and \textcolor{blue}{$N$=200} which have been inferred from real gene expression data, namely the microarray data extracted from metastatic melanoma cells studied in Bittner, {\it et al} \cite{bittner-2000-melanoma}. Microarrays provide the relative activity of a large number of genes within a sample. 
 
 Before presenting the results of the control method on these real PBNs, we briefly outline the network inference method \cite{CN2022-infer}. This builds on standard inference based on Coefficients of Determination (CODs) \cite{Kim-2000-CODs}, which describe how good a function $f(X)=Y$ performs when predicting $Y$ using $X$. Here, $Y$ is \textcolor{blue}{the} target gene (with state 1 or 0), and $X$ is the set of genes that have a directed edge into $Y$ (\textcolor{blue}{$inp^Y$}). Typically, a linear predictor is used:
  \begin{equation}
     f(X) = a_1X_1 \cdot a_2X_2 \dots a_nX_n + b
  \end{equation}
 COD\textcolor{blue}{s} are used to measure the effect on the prediction of adding a specific gene $x \in (N \setminus Y)$ to the set $X$, and they take the form
 \begin{equation}
     \theta = \frac{e_X - e_{X \bigcup x}}{e_X}
 \end{equation}
where $e_X$ is the error of the model when predicting $Y$ given $X$, and $e_{X \bigcup x}$ is the error of the model when predicting $Y$ given $X$ {\it and} some additional gene $x$.

However, there exist non-linear relationships in the data, hence linear predictors are not sufficient in this context, i.e., if the inputs for $Y$ are $x_1$ and $x_2$, and $Y$ is activated (state=1) when $x_1 \neq x_2$, then the linear predictor would entirely fail inferring this causal relationship. For this reason, we infer the functions straight from the single cell RNAseq data, based on \cite{Kim-2002-markov} - once the inputs are determined, for each possible input combination we note the rate of the output being True. 

This results in a Lookup table (LUT), which holds the rates of the output being True for each input combination. It can be used as a function.
To infer such functions from data, given a particular set of input data $X$ and output data $Y$, for each observed input combination, $x$, the probability $p$ is given by:
\begin{equation}
    p_x = P(Y=1 | X=x) = \frac{\#(Y=1|X=x)}{\#(X=x)}
\end{equation}
where $\#(Z)$ denotes the number of instances where $Z$ is True.

In terms of pre-processing in the data set comprising 8,150 genes and 31 samples, \textcolor{blue}{for gene selection discriminative weights are computed, which determine how a gene changes during the experiment compared to the control cells.} Since gene expression data shows the relative activation levels of genes within cells, the values of which are between 0 and $\infty$, the gene data has to be quantized, and there are various methods for this \cite{Mussel-2016-binarise-discont-package}. \textcolor{blue}{Here, the thresholds were computed by} applying k-means clustering, drawing upon Shmulevich \& Dougherty in \cite{Shmulevich-2010-book}.
This results in quantized tuples $(F,m)$ where $F$ is the set of functions \textcolor{blue}{with corresponding probabilities} (e.g., Table \ref{table:pbfuncsn10}), and $m$ the input mask. 

Next, we present results of the proposed control method on large PBNs from real gene data used in cancer biology. 

\textbf{Melanoma PBN $N$=28. }  We consider a PBN of size 28 from the same metastatic melanoma gene data \cite{bittner-2000-melanoma}. We use the exact same genes also studied in \cite{sirin-2013-BatchRL}, for the purpose of direct comparison. Intervention in this case \textcolor{blue}{concerns perturbations on} the {\it pirin} gene, and the objective is to arrive at a subset of network states where another gene, namely WNT5A, is OFF. The state space (just under 270 million states) makes calculating attractors prohibitive. Like \cite{par-2006-pbn7, sirin-2013-BatchRL}, we compute the steady state distribution (SSD) instead. To control such a network we aim to increase the mass probability of the network ending up in a desirable state.

\begin{figure}[h]
    \begin{minipage}{.5\textwidth}
        \centering
    	\includegraphics[width = 0.9\linewidth]{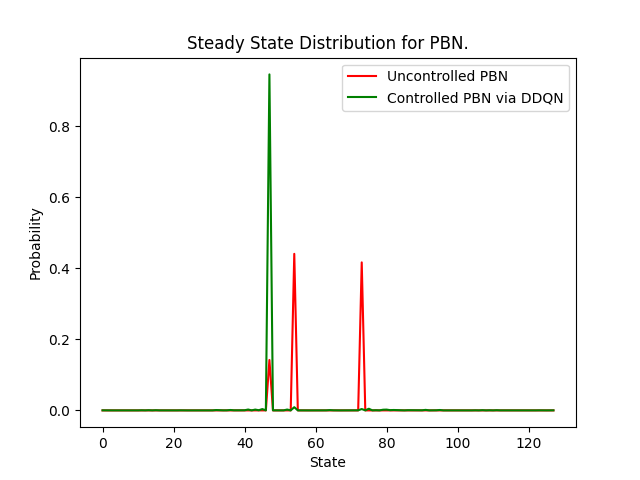}
    	\caption{States left of 64 are desirable; the rest are undesirable.}
    	\label{fig:SSD7BRL}
    \end{minipage}
	\begin{minipage}{.5\textwidth}
	    \centering
    	\includegraphics[width = 0.9\linewidth]{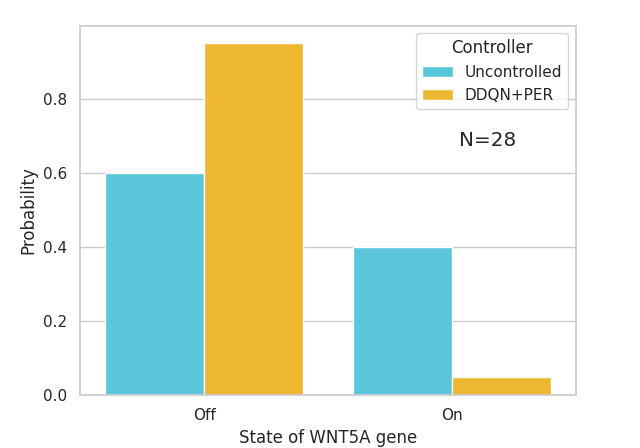}
    	\caption{Increase in probability mass of arriving at desirable state (PBN $N$=28).}
    	\label{fig:PBN28-success}
	\end{minipage}
\end{figure}

We have also computed the steady state distribution histogram for this melanoma network but the $N$=28 has $2^{28}$ states on the x-axis making straightforward visualisation rather challenging. Instead, we show the SSD histogram for the Melanoma $N$=7 in Fig. \ref{fig:SSD7BRL} which comprises $2^7=128$ states on the x-axis. We note it agrees with the SSD in \cite{sirin-2013-BatchRL}; there are three prominent states, two of which are desirable. However, where the control of \cite{sirin-2013-BatchRL} increase most of the prominent states uniformly, our method only amplifies the desired attractor state out of the three prominent states.

For the \textcolor{blue}{stabilization} of the melanoma PBN $N$=28, the output layer of the \textcolor{blue}{DDQN} is of size 2; one representing the Q value of taking or not taking an action, with an action defined as flipping the state of the \textit{pirin} gene.

\textcolor{blue}{The RL environment starts approaching the state space scale of environments tested in the original DQN paper \cite{mnih-DeepQ-Learning} and thus we adopt its default parameters for the experiments. The exploration fraction was set to $0.1$, the \textit{target} Q network is updated every 10,000 training steps and we train for 150,000. The primary adjustment is the use of $256$ as the batch size.}

The agent achieves $95.1\%$ control, see Fig. \ref{fig:PBN28-success}, which is a significant improvement to the $80\%$ \textcolor{blue}{reported previously in the literature, see \cite{sirin-2013-BatchRL}}. 

\textbf{Melanoma PBN $N$=70.} The proposed DDQN with PER control method was tested against a larger PBN consisting of 70 nodes. This network was generated from the same data set on metastatic melanoma provided by Bittner, {\it et  al} \cite{bittner-2000-melanoma}. 

The selection of genes for this PBN comprises the 7 found in the studies of  \cite{sirin-2013-BatchRL,par-2006-pbn7} discussed earlier in the context of the smaller Melanoma PBNs, and the rest were appended from the gene list in Bittner, {\it et al} \cite{bittner-2000-melanoma} based on their discriminative weights, which were computed as described in \cite{bittner-2000-melanoma}.

\begin{figure}[h]
    \begin{minipage}{.5\textwidth}
        \centering
        \includegraphics[width = 0.9\linewidth]{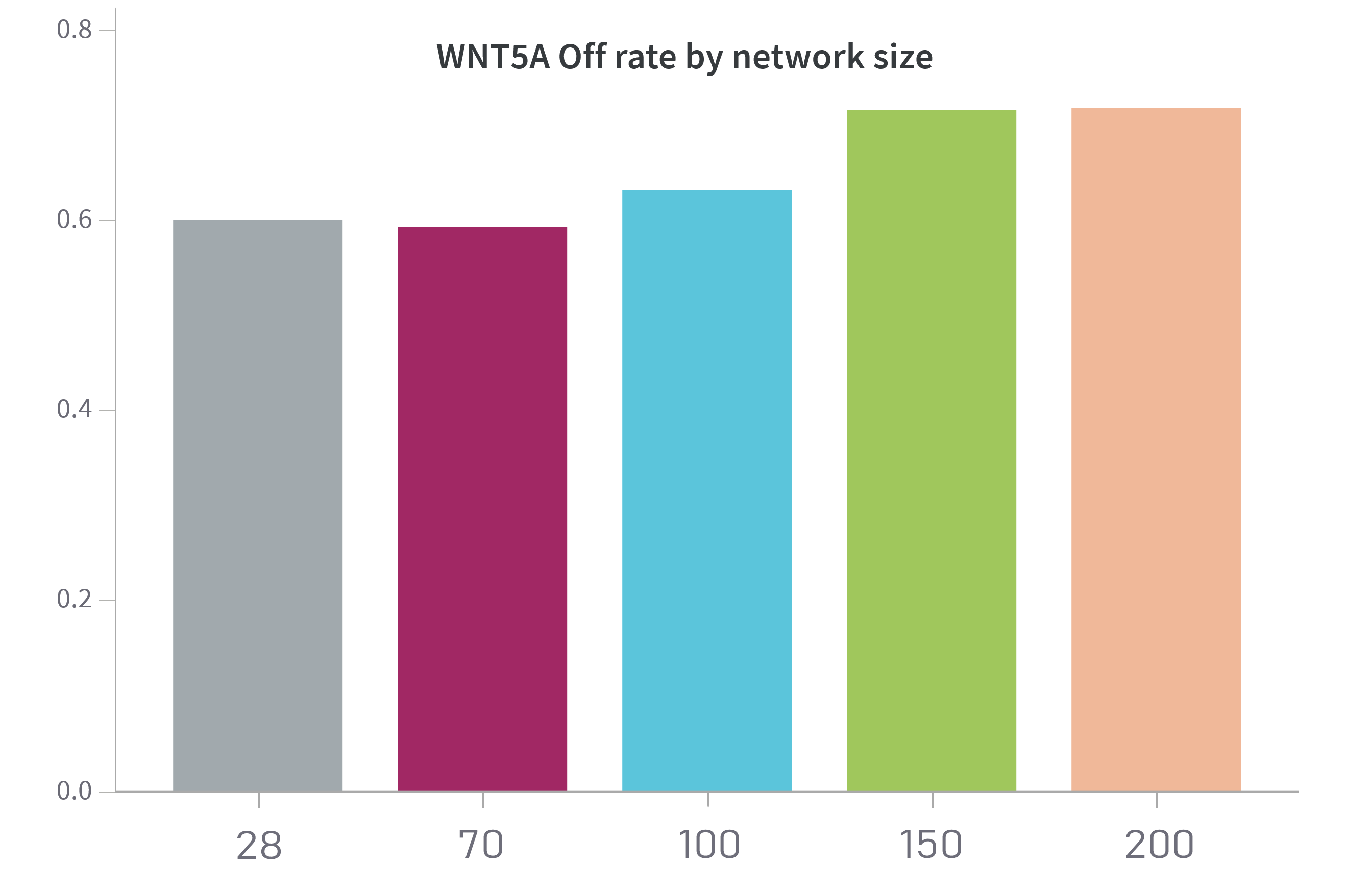}
    	\caption{Steady state distribution for different size PBNs from Bittner's Melanoma gene data \cite{bittner-2000-melanoma}.}
    	\label{fig:comparative-ssds}
    \end{minipage}
    \begin{minipage}{.48\textwidth}
        \centering
        \includegraphics[width = 0.9\linewidth]{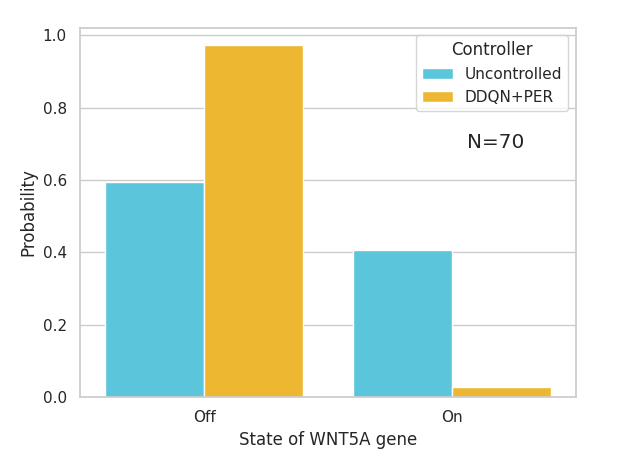}
    	\caption{Increase in probability mass of arriving at desired state (PBN $N$=70).}
    	\label{fig:cumhistPBN70}
    \end{minipage}
\end{figure}

\textcolor{blue}{The SSD for this network agrees with $N$=28, see Fig. \ref{fig:comparative-ssds}, which in turn agrees with the literature, as discussed before. }

The settings for training the DDQN with PER agent were adjusted so that the MLP architecture of the DQN has a larger amount of neurons in the first hidden layer, resulting in a (128 $\rightarrow$ 64) hidden layer structure.

Additionally, we adjust the batch size back down to $128$ and decrease the buffer size to 5,120, as well as update the target network every 1,000 training steps. The exploration fraction was also adjusted to $0.5$. 

The agent achieves $97.3\%$ successful control in this large PBN \textcolor{blue}{despite operating in a most challenging environment comprising $2^{70}$ states,} see Fig. \ref{fig:cumhistPBN70}. 

\textcolor{blue}{\textbf{Melanoma PBN $N$=200.} In order to truly test the scalability of the approach, we then generated a network of over double the size weighing in at $N$=200 nodes. The SSD for this much larger network does in fact agree with that of the $N$=28 and $N$=70, although it has drifted up slightly, which is in line with the trend observed in SSDs of networks of intermediate size generated from the same data as well (see Fig. \ref{fig:comparative-ssds}).}

\textcolor{blue}{The settings for training the RL agent to operate in the $2^{200}$ state space of this PBN were based on those of the melanoma PBN $N$=28 with some changes: the exploration fraction was adjusted to $0.5$ to afford more exploration steps and the MLP architecture extended to 3 layers (256 $\rightarrow$ 128 $\rightarrow$ 64 architecture) to accommodate the 200 input size.} 

\textcolor{blue}{\begin{figure}[h]
	\centering
	\scalebox{1}{\includegraphics[width = 0.4\linewidth]{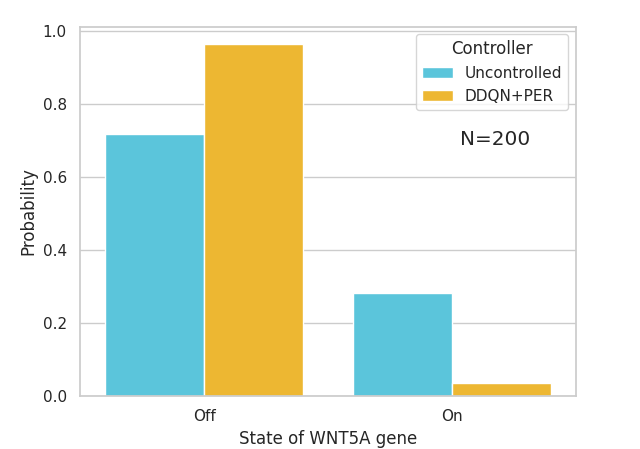}}
	\caption{Increase in probability mass of arriving at desired state (PBN $N$=200).}
	\label{fig:cumhistPBN200}
\end{figure}
}

\textcolor{blue}{When this large PBN is controlled by the RL agent, it spends $96.4\%$ of the time in the desirable states,
see Fig. \ref{fig:cumhistPBN200}.
To the best of our knowledge, this is the largest PBN to be controlled in the literature to date.}

\subsection{Comparison \& Discussion}
\label{subsec:discuss}

The model-free, deep reinforcement learning (DDQN with PER) approach to stabilization of PBNs has been tried on a number of synthetic and real PBNs generated directly from gene expression data.
On the synthetic PBN $N$=20, which comprises 1,048,576 states, the method achieves stabilization to an attractor that occurs 1 out of 10,000 times, within a horizon of 11.  \textcolor{blue}{The method achieves stabilization 99\% of the time} when allowed up to 15 interventions.

On the well known Melanoma PBN $N$=7 \textcolor{blue}{ \cite{par-2006-pbn7} and \cite{sirin-2013-BatchRL},} the method produces a policy for stabilization at a rate of 99.72\%. This compares favourably to 96\% in \cite{par-2006-pbn7} and 98\% in \cite{sirin-2013-BatchRL}.  

\textcolor{blue}{On significantly larger networks, we have demonstrated over 96.4\% successful control on the Melanoma PBN $N$=70 and PBN $N$=200 generated in the same way from the same gene data provided by Bittner, {\it et al} \cite{bittner-2000-melanoma}}. By means of direct comparison, on the Melanoma PBN $N$=28 which is also studied in \cite{sirin-2013-BatchRL}, we get 95.1\% successful control compared to 80\% (see Sec. 4.2 in \cite{sirin-2013-BatchRL}) hence, a significant improvement.

\textcolor{blue}{The approach in \cite{Acernese-2020-DDQN} applies Deep RL (in fact, the DDQN with PER method originally proposed by Papagiannis \& Moschoyiannis in \cite{papagiannis-2019-drl-rbn,CN2020-drl-pbn} (which is also adapted here), to the output tracking problem in Boolean networks. 
It is applied only to situations where control inputs are available (hence, PB{\it C}Ns) and the target domain is an attractor. In contrast, we presented an integrative DRL framework that can address the full set of nodes as potential control nodes. In addition, the target domain in our study can be an attractor (fixed point or cyclic) {\it but also} a pre-assigned subset of the state space; in the latter case we have shown how we validate the control method using the corresponding steady state distribution before and after control is applied (recall Fig. \ref{fig:PBN28-success}, \ref{fig:cumhistPBN70}, \ref{fig:cumhistPBN200}). Finally, for the $N$=28 the approach in \cite{Acernese-2020-DDQN} requires $2 \times 10^6 =$ 2M (million) episodes in training 
while, indicatively, our approach on a (different) $N$=28 network requires only 150,000 time steps.
Finally, we demonstrated the scalability of the approach here by showing successful control of Melanoma PBN with $N=200$. }

\begin{table}[h]
\begin{center}
\scalebox{0.8}{\begin{tabular}{|c|c c c c|}
         Network & DDQN+PER & DDQN & PPO & TRPO\\
		 \hline
		 N=28 & \textbf{0.951} (std: 0.028) & 0.910 (std: 0.045) & \textbf{0.951} (std: 0.008) & 0.947 (std: 0.013) \\
		 N=70 & \textbf{0.973} (std: 0.125) & 0.812 (std: 0.071) & 0.971 (std: 0.008) & 0.970 (std: 0.002) \\
		 N=200 & \textbf{0.964} (std: 0.020) & 0.912 (std: 0.016) & 0.946 (std: 0.006) & 0.956 (std: 0.001) \\
	\end{tabular}}
	\caption{\label{table:alg_comparison} \textcolor{blue}{Comparison of final probability mass of arriving at desired state by applying different RL algorithms for control, using the best run for each \& standard deviation from 5 runs}}
\end{center}	
\end{table}
\textcolor{blue}{Q-learning is a fundamental approach to RL but we also experimented with other state-of-the-art model-free RL algorithms, such as PPO \cite{schulman2017proximal} and its predecessor, TRPO \cite{schulman2015trust}. Table \ref{table:alg_comparison} summarises our experiments that show there is little improvement to be had. It is worth noting that PPO and TRPO demonstrate much lower standard deviation in final performance across multiple training runs compared to DDQN+PER. This is to be expected as those policy gradient methods are revered for their stability, but the best seed for DDQN+PER consistently outperformed them, albeit marginally, for the networks we tested.}

\section{Conclusion}
\label{sec:concl}

\textcolor{blue}{We have demonstrated that the  model-free, deep RL method (DDQN with PER) is successful in (set) stabilization of large-scale PBNs,  of up to 200 nodes, with  significant improvements in performance on Melanoma PBNs studied in the existing literature.} The method does not assume knowledge of the Probability Transition Matrix \textcolor{blue}{and does not require control input nodes to be known, although it can readily work with them if they are available}.

\textcolor{blue}{The time complexity, given by $\mathcal{O}(t ^. B ^. |\theta|)$ (Section \ref{sec:ddqn-control-pbn}) is {\it linearly} dependent on the time steps $t$ needed in training, for learning the optimal policy in reaching the target domain.}

\textcolor{blue}{In our experiments where the goal is to reach a specific attractor, this is chosen to be the one least likely to naturally occur. The intervention horizon was set to 0.8-1.5\% of the random perturbations (on average) required to reach the target domain. The PBNs (and the PB{\it C}Ns) we addressed are highly stochastic; indicatively, the PBN $N$=10, has 1,296 possible network realisations at each step, compared to a handful of possible PBN realisations considered in existing works. Set stabilizaton of a real Melanoma PBN $N$=200 was achieved and was demonstrated via the favourable shift in the corresponding SSD towards the target domain.}

Possible extensions include targeting known pitfalls in RL such as the need for frequent observations of the network state, as these can be computationally costly. Work is under way in this direction, on exploring sampled-data control, effectively varying the window of deciding on the next perturbation by considering {\it options} in RL \textcolor{blue}{within a semi-MDP framework \cite{Chatzaroulas-SDC-DRL-2022}}. 

One other interesting extension would be to determine ways to utilize the PBN's transition patterns observed during training to improve the learning efficiency. To this end, {\it control} nodes, in the sense of \cite{liu-2011-controllability,moschoyiannis-2016-control,moschoyiannis-cna-2017,Shmulevich-2017-IMP}, may be useful. The identification of pinning control nodes in \cite{Lu-2021-acyclic} based on local neighbours rather than global state information could possibly be leveraged in this respect. 

\textcolor{blue}{
Alongside a similar vein of research would be to explore the application of \textit{model-based} RL approaches \cite{kurutach2018model,janner2019trust} on real cells directly and infer the model ``on-the-go'' to apply RL on it. This could lead to increased real-word gene data sample efficiency for deriving a control policy.}

\textcolor{blue}{
Finally, an exciting frontier for RL methods at the moment is using sequential models to learn a policy. Exploring the application of Decision Transformers \cite{chen2021decision} to the analysis of large-scale PBNs seems worthwhile.}

Code and example networks can be found at:\\ \textcolor{blue}{\url{https://github.com/UoS-PLCCN/pbn-rl/} and\\ \url{https://github.com/UoS-PLCCN/gym-PBN}}

\bibliographystyle{unsrt}
\bibliography{drl-pbn}

\clearpage
\appendix{Hyperparameter Settings}
\begin{figure}[h]
	\centering
	\scalebox{1}{\includegraphics[width = 0.7\linewidth]{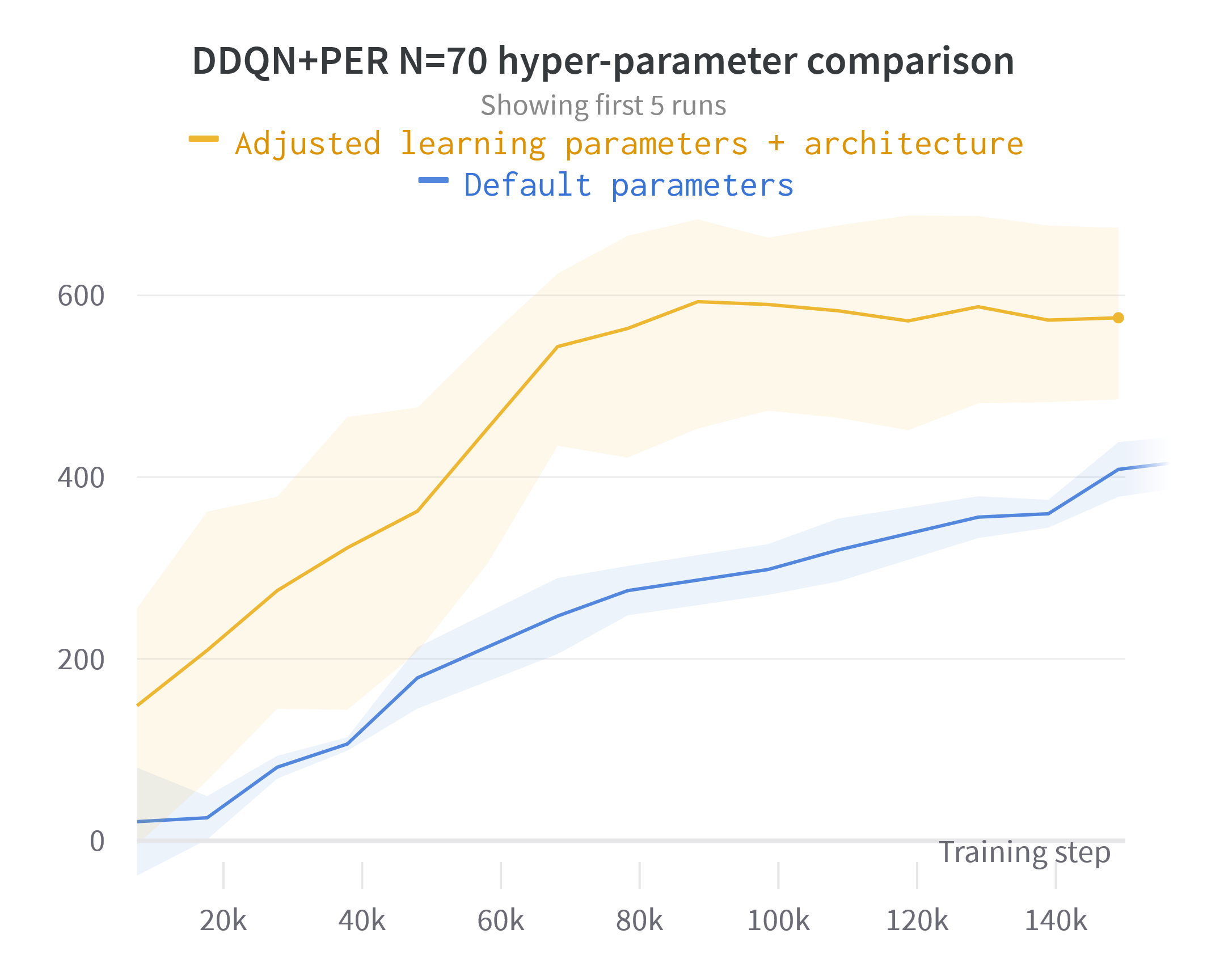}}
	\caption{Comparison of different DDQN+PER parameter settings for the $N$=70 network.}
	\label{fig:n70_ablation}
\end{figure}

\begin{figure}[h]
	\centering
	\scalebox{1}{\includegraphics[width = 0.7\linewidth]{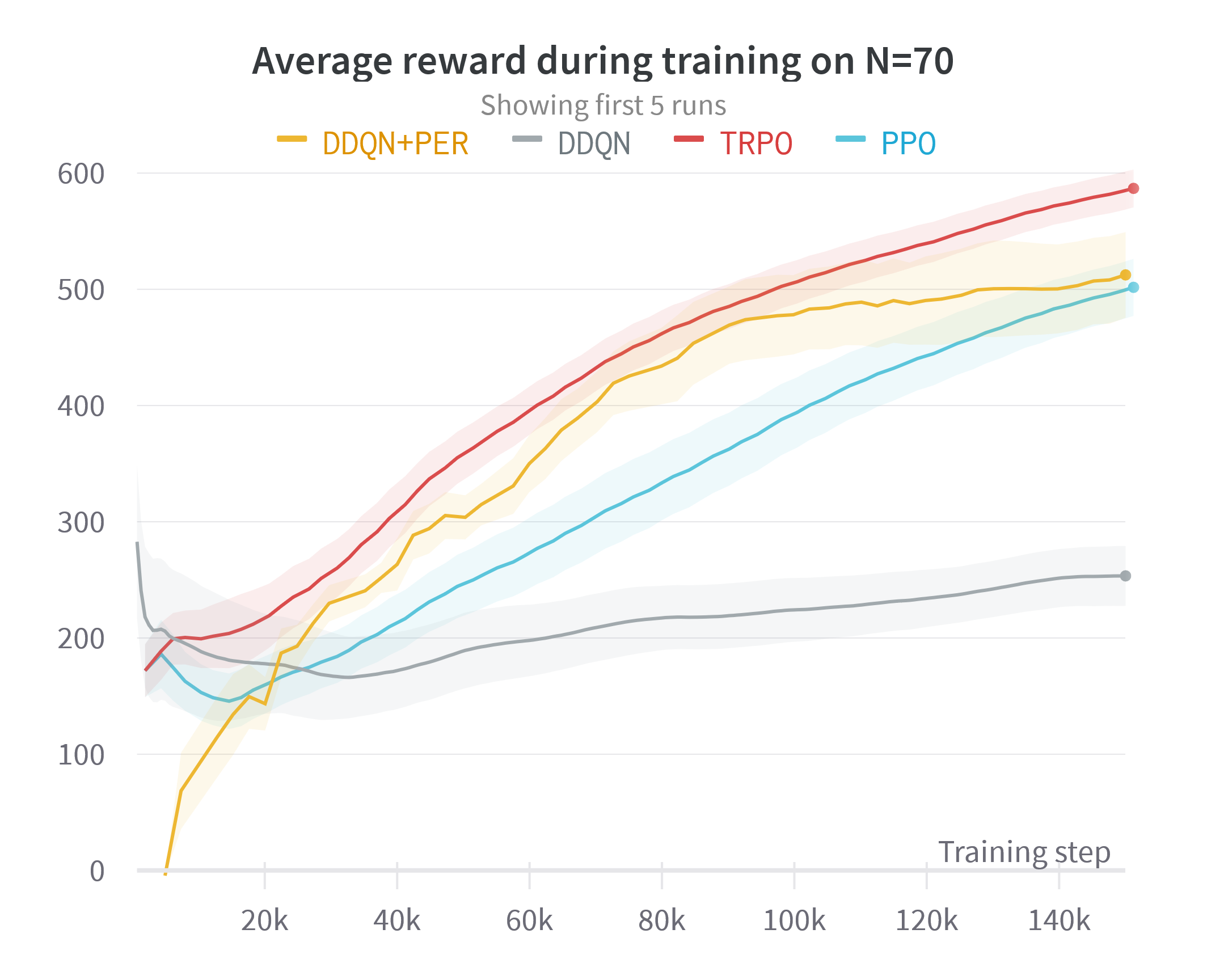}}
	\caption{Comparison of learning performance between the model-free methods on the $N$=70 PBN.}
	\label{fig:n70_train}
\end{figure}

Our hyperparameter optimization process was to start with the parameters recommended in one of the breakout papers that brought the potential of DQN to light\footnote{V. Mnih and et al, “Human-level control through deep reinforcement learning,” Nature, vol. 518, no. 7540, pp. 529–533, 2015.}. It is worth noting that this set of parameters is also the set of default parameters for the popular Reinforcement Learning library Stable baselines3\footnote{\url{https://github.com/DLR-RM/stable-baselines3}}. This set of hyperparameters worked out of the box for the $N$=28 Melanoma PBN, but that was not the case  for $N$=70 and $N$=200, showing limited to no learning signal.

For the $N$=70, we first tried adjusting the ``learning'' parameters--as in parameters that directly affect how and when the agent learns. These include the batch size, the learning rate, the interval at which the \textit{target} Q network gets updated with the latest parameters and the buffer size. DDQN is an off-policy model-free RL algorithm, but the on-policy baselines we tried in the case of PPO \footnote{J. Schulman, F. Wolski, P. Dhariwal, A. Radford, and O. Klimov, “Proximal policy optimization algorithms,” arXiv preprint arXiv:1707.06347,2017.} and TRPO\footnote{J. Schulman, S. Levine, P. Abbeel, M. Jordan, and P. Moritz, “Trust region policy optimization,” in International conference on machine learning. PMLR, 2015, pp. 1889–1897} did not struggle in this network whatsoever. Thus, we adjusted the DDQN parameters to steer its behaviour closer to an on-policy algorithm. By dramatically reducing the buffer size from 1,000,000 to 5,120, we quickly discard old samples, although not as quickly as on-policy methods. To avoid overfitting on the much smaller dataset, we reduce the batch size from $256$--which we used for the $N$=28 PBN--to $128$. Finally, observing that the dimension of the input layer ($70$ neurons) was smaller than that of the first hidden layer ($64$ neurons), we adjusted the first hidden layer to $128$ neurons. Finally, we increased the exploration fraction to $0.5$ in order to afford more exploration, and copied over the parameters every 1,000 time steps instead of the default of 10,000.

This set of parameters performed significantly better than the default set of parameters, as shown in Fig. \ref{fig:n70_ablation}, and much closer to the well-performing baselines, as shown in Fig. \ref{fig:n70_train}.

\begin{figure}[h]
	\centering
	\scalebox{1}{\includegraphics[width = 0.7\linewidth]{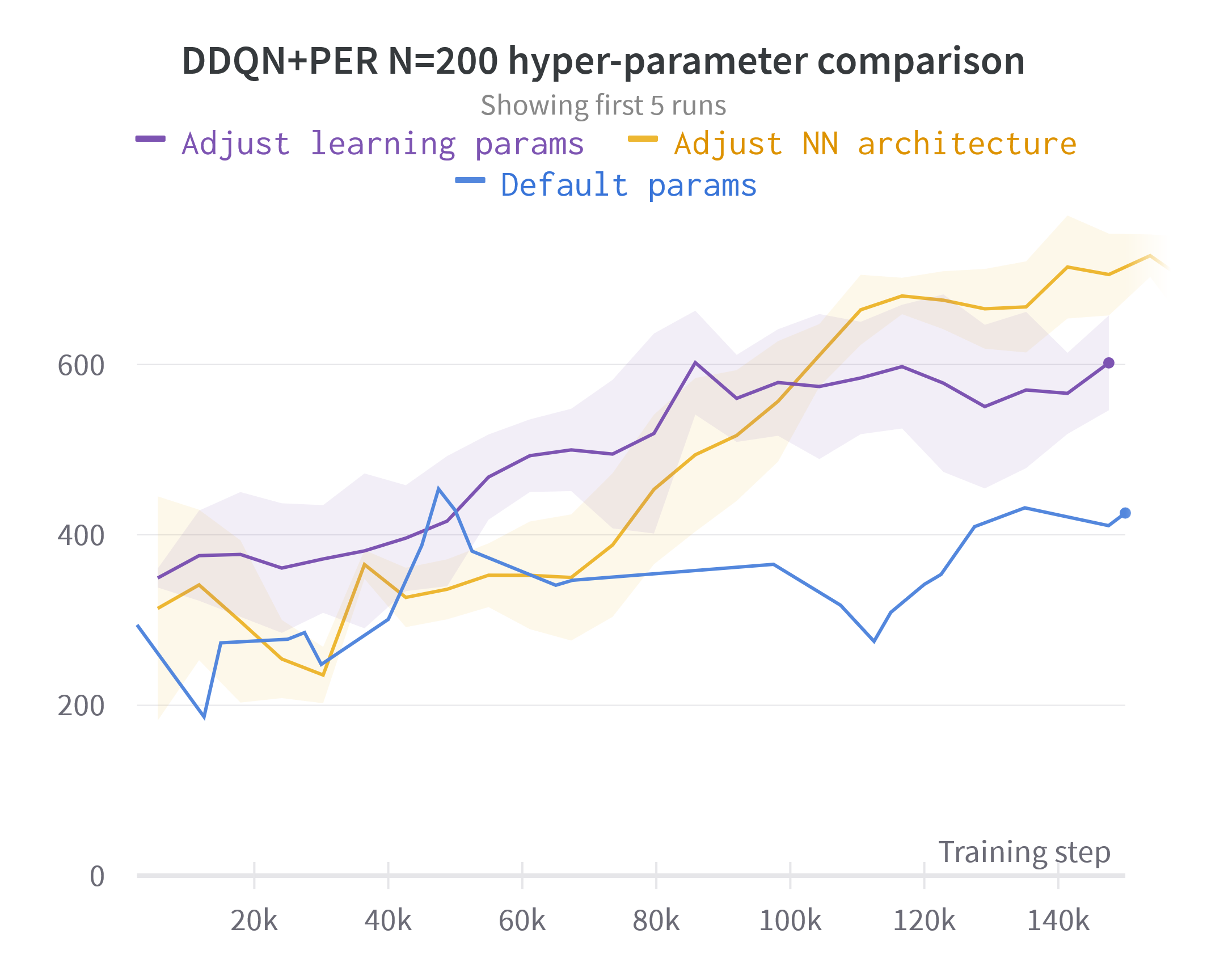}}
	\caption{Comparison of different DDQN+PER parameter settings for the $N$=200 network.}
	\label{fig:n200_ablation}
\end{figure}

\begin{figure}[h]
	\centering
	\scalebox{1}{\includegraphics[width = 0.7\linewidth]{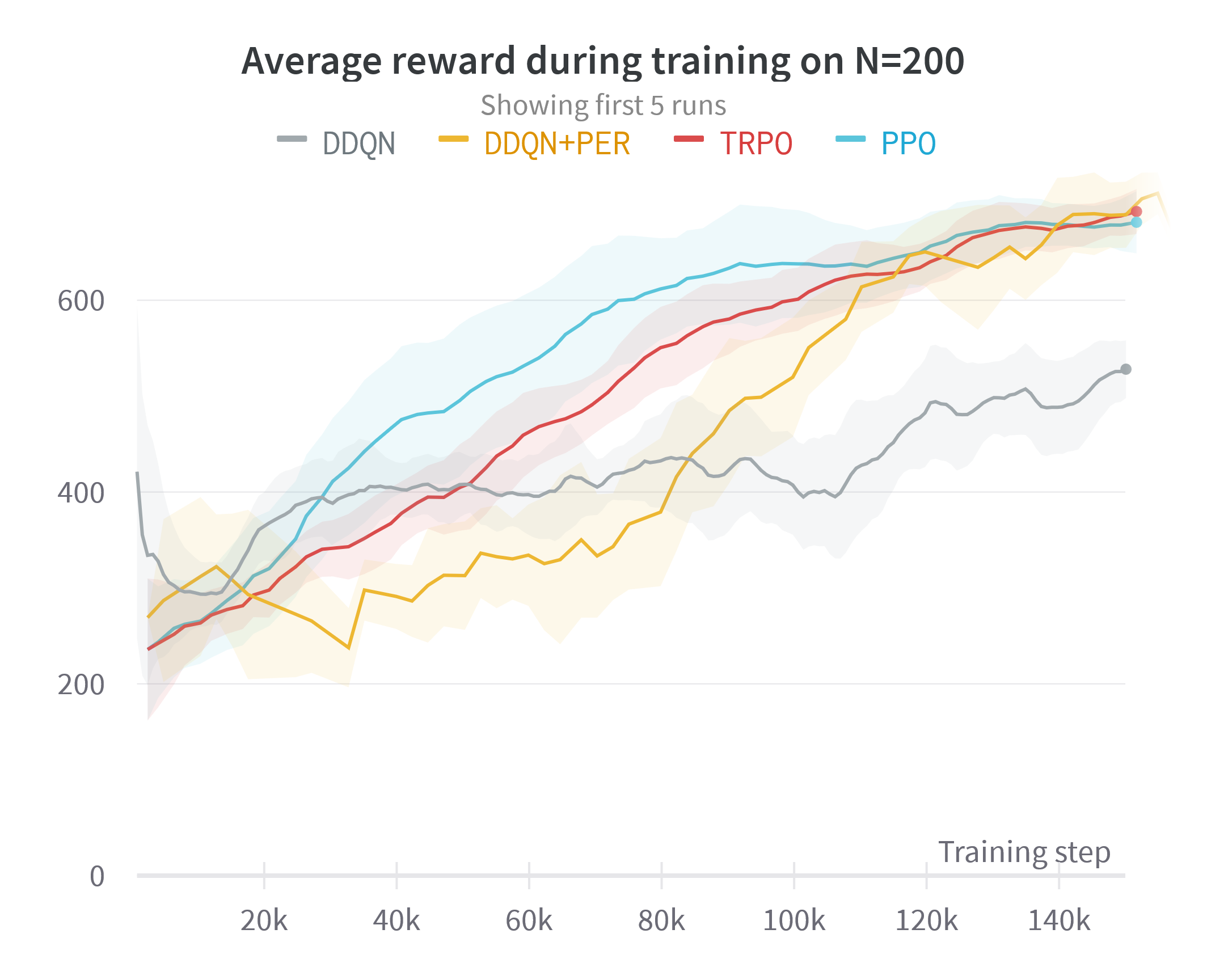}}
	\caption{Comparison of learning performance between the model-free methods on the $N$=200 PBN.}
	\label{fig:n200_train}
\end{figure}

For the $N$=200, we initially began with applying all the learning parameter changes from $N$=70. While this did yield an increase in performance, it was still not up to par with the rest of the baselines. We then tried adjusting the MLP architecture itself in a similar way we did to the $N$=70, except this time add a third hidden layer at the front of $256$ neurons to accommodate the 200 inputs. Additionally, we adjusted the buffer size back up to its default value of 1,000,000 and batch size back to $256$ as well as the target Q network parameter copy interval to 10,000, motivated by the much larger state space and the fact that the default--more ``off-policy''--parameters still showed \textit{some} learning signal, unlike the $N$=70 where they showed almost none at all. This final set of parameters outperformed the defaults and the lightly tuned $N$=70 parameters as shown in Fig. \ref{fig:n200_ablation}, and once again performed similar to PPO and TRPO as shown in \ref{fig:n200_train}.

Overall, we suggest starting with the default parameters from \footnote{V. Mnih and et al, “Human-level control through deep reinforcement learning,” Nature, vol. 518, no. 7540, pp. 529–533, 2015.} and adjusting in the following order:

\begin{enumerate}
    \item \textbf{MLP Architecture} - In both cases, having an architecture that better fit the input and output sizes helped dramatically.
    \item \textbf{Buffer size} - 1,000,000 is extremely large especially when PER is used. While it might not lead to bad performance in a lot of cases, sometimes reducing this can speed up learning as older experiences are discarded faster since they get wiped out of the buffer due to its size.
    \item \textbf{Target network update interval} - Lowering this is worth trying alongside making the buffer size smaller as in some cases the target network does not update quickly enough at 10,000 to facilitate Q learning.
    \item \textbf{Batch size / learning rate} - These do not seem to affect performance when kept within reasonable settings, but it is worth noting that oftentimes $256$ batch size might be too large.
\end{enumerate}

\end{document}